\pdfoutput=1
\documentclass[11pt]{article}

\usepackage[]{acl}
\usepackage{times}
\usepackage{latexsym}

\usepackage[T1]{fontenc}
\usepackage[utf8]{inputenc}

\usepackage{microtype}

\usepackage{inconsolata}
\usepackage{enumitem}
\usepackage{tabularx}
\usepackage{bm}
\usepackage{amsfonts}
\usepackage{amssymb}
\usepackage{amsmath}
\usepackage{pifont}% http://ctan.org/pkg/pifont
\usepackage{bbm}
\usepackage{svg}
\usepackage{moresize}
\usepackage{caption}
\usepackage{subcaption}
\usepackage{graphicx}
\usepackage{enumitem}
\usepackage{multirow}
\usepackage{multicol}
\usepackage[para]{threeparttable}
\usepackage{booktabs}
\usepackage{hyperref}
\usepackage{array}
\usepackage{colortbl}   
\usepackage{xspace}
\usepackage[normalem]{ulem}
\useunder{\uline}{\ul}{}
\usepackage{pgffor}
\usepackage{longtable}
\usepackage{graphicx}
\usepackage{cleveref}
\usepackage{makecell}
\usepackage{csquotes}
\usepackage{mathtools}
\usepackage{kotex}

\usepackage{xparse}
\usepackage[strict]{changepage}

\makeatletter
\def\gcmidrule{\arrayrulecolor{gray!20}% Switch colour to lightgray
    \noalign{\ifnum0=`}\fi
    \@ifnextchar[{\@gcmidrule}{\@gcmidrule[\cmidrulewidth]}}
\def\@gcmidrule[#1]{\@ifnextchar({\@@gcmidrule[#1]}{\@@gcmidrule[#1]()}}
\def\@@gcmidrule[#1](#2)#3{\@@@gcmidrule[#3]{#1}{#2}}
\def\@@@gcmidrule[#1-#2]#3#4{\global\@cmidla#1\relax
    \global\advance\@cmidla\m@ne
    \ifnum\@cmidla>0\global\let\@gtempa\@cmidrulea\else
    \global\let\@gtempa\@cmidruleb\fi
    \global\@cmidlb#2\relax
    \global\advance\@cmidlb-\@cmidla
    \global\@thisrulewidth=#3
    \@setrulekerning{#4}
    \ifnum\@lastruleclass=\z@\vskip \aboverulesep\fi
    \ifnum0=`{\fi}\@gtempa
    \noalign{\ifnum0=`}\fi\futurenonspacelet\@tempa\@xgcmidrule}
\def\@xgcmidrule{%
   \ifx\@tempa\gcmidrule
       \vskip-\@thisrulewidth
       \global\@lastruleclass=\@ne
   \else \ifx\@tempa\morecmidrules
       \vskip \cmidrulesep
       \global\@lastruleclass=\@ne\else
       \vskip \belowrulesep
       \global\@lastruleclass=\z@
   \fi\fi
   \ifnum0=`{\fi}
  \arrayrulecolor{black}}% Switch colour back to black
\makeatother

\DeclareMathOperator{\catop}{type}
\DeclareMathOperator{\softmax}{softmax}

\DeclareMathOperator{\idx}{index}

\DeclareMathOperator{\layernorm}{LayerNorm}

\newcommand{\xmetricop}{\textsc{EntRec}}
\newcommand{\ymetricop}{\textsc{CnstScore}}

\newcommand{\dims}[1]{\ensuremath{\mathbb{R}^{#1}}}

\newcommand{\cpromptshort}{\ensuremath{\tau_\text{2H}}}
\newcommand{\mpromptshort}{\ensuremath{\tau_\text{1H}}}

\newcommand{\cprompt}{\ensuremath{\textify{\outxy}}}

\newcommand{\mprompt}{\ensuremath{\textify{\outy}}}

\newcommand{\symrel}{\ensuremath{r}}
\newcommand{\syment}{\ensuremath{e}}
\newcommand{\symentx}{\ensuremath{\syment_1}}
\newcommand{\symenty}{\ensuremath{\syment_2}}
\newcommand{\symentz}{\ensuremath{\syment_3}}

\newcommand{\symrelx}{\ensuremath{\symrel_1}}
\newcommand{\symrely}{\ensuremath{\symrel_2}}

\newcommand{\eii}{\ensuremath{{e_{i+1}}}}
\newcommand{\ei}{\ensuremath{{e_i}}}
\newcommand{\ri}{\ensuremath{{r_i}}}

\newcommand{\entx}{\ensuremath{{\symentx}}}
\newcommand{\enty}{\ensuremath{{\symenty}}}
\newcommand{\entz}{\ensuremath{{\symentz}}}

\newcommand{\menty}{\ensuremath{{\symenty}}}

\newcommand{\mentyfirst}{\ensuremath{e_2^{\scalebox{0.7}[0.7]{(0)}}}}

\newcommand{\relx}{\ensuremath{\symrelx}}

\newcommand{\rely}{\ensuremath{\symrely}}

\newcommand{\symtemplate}{\ensuremath{t}}
\newcommand{\symmention}{\ensuremath{m}}
\newcommand{\templateforx}{\ensuremath{{\symmention}}}
\newcommand{\templatefory}{\ensuremath{{\symtemplate}}}

\newcommand{\out}[2]{\ensuremath{{#1}({#2})}}
\newcommand{\outx}{\ensuremath{\out{\relx}{\entx}}}
\newcommand{\outy}{\ensuremath{\out{\rely}{\enty}}}
\newcommand{\outxy}{\ensuremath{\out{\rely}{\out{\relx}{\entx}}}}

\newcommand{\templatefx}{\ensuremath{\ensuremath{\templateforx_{\relx}}}}
\newcommand{\templatefy}{\ensuremath{\ensuremath{\templatefory_{\rely}}}}

\newcommand{\templatex}[1]{\ensuremath{\ensuremath{\templatefx({#1})}}}
\newcommand{\templatey}[1]{\ensuremath{\ensuremath{\templatefy({#1})}}}

\newcommand{\cat}[1]{\ensuremath{\catop(#1)}}

\newcommand{\mentionize}[1]{\ensuremath{\mu({#1})}}
\newcommand{\mt}{\ensuremath{\mentionize{\outx})}}

\newcommand{\symtextify}{\ensuremath{\tau}}
\newcommand{\textify}[1]{\ensuremath{\symtextify({#1})}}

\newcommand{\datasetname}{\textsc{TwoHopFact}}

\newcommand{\xmetricreal}[1]{\ensuremath{\xmetricop^{#1}(\enty, \cpromptshort)}}
\newcommand{\ymetricreal}{\ensuremath{\ymetricop(\cpromptshort, \mpromptshort)}}

\newcommand{\xmetricfunc}[1]{\ensuremath{\xmetricop({#1})}}
\newcommand{\ymetricfunc}[1]{\ensuremath{\ymetricop({#1})}}

\newcommand{\hcolor}{violet}
\newcommand{\hhcolor}{brown}
\newcommand{\hlight}[1]{\textcolor{\hcolor}{{#1}}}
\newcommand{\hhlight}[1]{\textcolor{\hhcolor}{{#1}}}

\newcommand{\negcprompt}{\ensuremath{\tau'_\text{2H}}}
\newcommand{\negentx}{\ensuremath{e'_1}}
\newcommand{\negrelx}{\ensuremath{r'_1}}

\newcommand{\btype}{\catstr{\cat{\entx}'s \cat{\relx}}}
\newcommand{\ftype}{\catstr{\cat{\rely} of \cat{\entx}'s \cat{\relx}}}

\newcommand{\name}[1]{\ensuremath{n_{#1}}}

\newcommand{\appositiontarget}{\mentyfirst}

\newcommand{\residstream}[1]{\ensuremath{\mathbf{x}^{#1}}}

\newcommand{\newresidstream}[1]{\ensuremath{\mathbf{\hat{x}}^{#1}}}

\newcommand{\xmetricreparam}{\xmetricfunc{\residstream{l}}}
\newcommand{\ymetricreparam}{\ymetricfunc{\residstream{l}}}

\newcommand{\catstr}[1]{``{#1}''}

\title{Do Large Language Models Latently Perform Multi-Hop Reasoning?}

\author{
\vspace{5px}
Sohee Yang\textsuperscript{1,2} \quad Elena Gribovskaya\textsuperscript{1} \quad Nora Kassner\textsuperscript{1} \quad Mor Geva\textsuperscript{3,4$*$} \quad Sebastian Riedel\textsuperscript{1,2$*$} \\ 
 \vspace{5px}
\textsuperscript{1}Google DeepMind \quad \textsuperscript{2}UCL \quad \textsuperscript{3}Google Research \quad \textsuperscript{4}Tel Aviv University \\  \vspace{2px}
\normalsize{\texttt{\{soheeyang,egribovskaya,norakassner,pipek,srriedel\}@google.com}}
}

\begin{document}
\maketitle
\def\thefootnote{*}\footnotetext{Corresponding authors.}\def\thefootnote{\arabic{footnote}}
\begin{abstract}
We study whether Large Language Models (LLMs) latently perform multi-hop reasoning with complex prompts such as ``The mother of the singer of `Superstition' is''. We look for evidence of a latent reasoning pathway where an LLM (1) latently identifies ``the singer of `Superstition''' as Stevie Wonder, the \textit{bridge entity}, and (2) uses its knowledge of Stevie Wonder's mother to complete the prompt. We analyze these two hops individually and consider their co-occurrence as indicative of latent multi-hop reasoning. For the first hop, we test if changing the prompt to indirectly mention the bridge entity instead of any other entity increases the LLM's internal recall of the bridge entity. For the second hop, we test if increasing this recall causes the LLM to better utilize what it knows about the bridge entity. We find strong evidence of latent multi-hop reasoning for the prompts of certain relation types, with the reasoning pathway used in more than 80\% of the prompts. However, the utilization is highly contextual, varying across different types of prompts. Also, on average, the evidence for the second hop and the full multi-hop traversal is rather moderate and only substantial for the first hop. Moreover, we find a clear scaling trend with increasing model size for the first hop of reasoning but not for the second hop. Our experimental findings suggest potential challenges and opportunities for future development and applications of LLMs.\looseness-1\footnote{Our code and dataset are available at \url{https://github.com/google-deepmind/latent-multi-hop-reasoning} and \url{https://huggingface.co/datasets/soheeyang/TwoHopFact}, respectively.}
\end{abstract}

\section{Introduction}
\begin{figure}[t!]
\setlength{\belowcaptionskip}{-10px}
  \centering
  \includegraphics[width=0.4\textwidth]{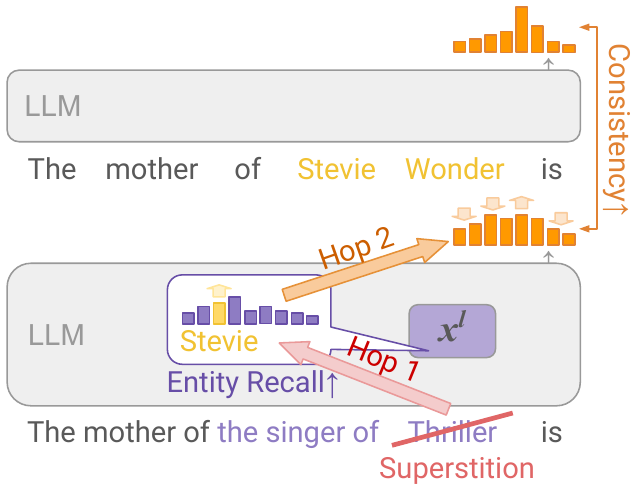}
  \caption{We investigate the latent multi-hop reasoning of LLMs. For the first hop, we change the input prompt to refer to the bridge entity (Stevie Wonder) and check how often it increases the model's internal recall of the bridge entity. For the second hop, we check if increasing this recall causes the model output to be more consistent with respect to what it knows about the bridge entity's attribute (mother of Stevie Wonder).}
  \label{fig:overview}
\end{figure}

Recent works have shown that Transformer-based~\citep{vaswani2017attention} Large Language Models (LLMs) store and retrieve factual information in their parameters to complete simple prompts such as \textit{``The mother of Stevie Wonder is''} ~\citep{Petroni2019-mu, Meng2022-br, Geva2020-jm, Geva2022-ah, Geva2023-lq, Zhu2023-nb}.
In addition, LLMs have demonstrated remarkable \textit{in-context} reasoning abilities when the necessary information is explicitly given as part of the input~\citep{Wei2022-qb}.
For example, models can infer ``Lula'' as a possible completion of \textit{``The mother of Stevie Wonder is Lula. The singer of `Superstition' is Stevie Wonder. The mother of the singer of `Superstition' is''}.
These findings raise a question: 
Do LLMs retrieve factual information stored in their parameters and perform \textit{latent multi-hop reasoning} when the information to reason from is \textit{not} given as a part of the input? 
For instance, when LLMs process the two-hop prompt \textit{``The mother of the singer of `Superstition' is''}, do they (1) figure out that ``the singer of `Superstition''' refers to Stevie Wonder and (2) use their knowledge of who Stevie Wonder's mother is to complete the prompt?

Answering this question is important. Evidence for such latent multi-hop reasoning would suggest that the LLM can \textit{connect and traverse through implicit knowledge} stored in their parameters rather than only storing information redundantly in its parameters. Future work could strengthen such paths of traversal, ultimately leading to more parameter-efficient and controllable models. Conversely, a lack of evidence would indicate more fundamental limitations of the Transformer architecture or training. It would also have critical implications for model editing: if complex facts are recalled instead of inferred, editing only base facts will never be enough since the changes cannot propagate~\citep{Onoe2023-wm,Zhong_undated-dp,Cohen2023-iq}.

In this work, we limit ourselves to prompts that express a composition of two facts such as ``\textit{The mother of the singer of `Superstition' is}'' that humans can complete with two hops by (1) inferring a \emph{bridge entity} (e.g., Stevie Wonder) and (2) inferring an attribute of that entity (e.g., who his mother is).
Then, we study how often LLMs process the prompt using a similar latent two-hop reasoning pathway, although this pathway may not be the most salient pathway that largely determines the predicted output.
To this end, we first study these hops individually, as shown in Figure~\ref{fig:overview}.
To study the first hop, we propose the \textit{entity recall score} to approximate LLM's internal recall of the bridge entity by projecting specific hidden representations to vocabulary space.
We test how changes to the input prompt affect this score. To study the second hop, we propose to measure the \textit{consistency score} between the distributions for completions of the two-hop prompt and an equivalent recall-based one-hop prompt (e.g., ``\textit{The mother of Stevie Wonder is}''). We check how often an intervention to increase the entity recall score increases consistency as an indication of second-hop utilization. Finally, we investigate how frequently both steps coincide.

To study latent two-hop reasoning with diverse types of fact composition, we introduce \datasetname{} dataset, which is based on Wikidata~\citep{vrandevcic2014wikidata} and consists of 45,595 two-hop prompts of 52 types of fact composition. We experiment with LLaMA-2~\citep{Touvron2023-jd} 7B, 13B, and 70B. Our findings can be summarized as follows. Across a wide range of fact composition types for the two-hop prompts, we find substantial evidence for the first hop of the multi-hop reasoning.
In about 70\% of the times where we change the prompt to indirectly mention the bridge entity, the later layers of the transformer show increased bridge entity recall.
For the second hop and overall traversal, the evidence appears weaker: in 60\% of the cases where we increase entity recall score, consistency goes up. Likewise, in about 40\% of the time, both hops work together~(compared to a random 25\% baseline); changing the descriptive mention increases the entity recall score, and increasing this recall score increases consistency. 

While the above aggregate statistics do not suggest a very prevalent use of the latent multi-hop reasoning pathway, it is worth pointing out that up to 23\% of the fact composition types demonstrate strong evidence of latent multi-hop reasoning, occurring in more than 80\% of the cases. This suggests that the pathway \emph{exists} but is highly contextual. Additionally, we focus on a very narrow interpretation of the pathway -- in reality, we expect it to be more distributed across layers and tokens. Hence, the effects we see might be a lower bound on the model's ability to perform latent two-hop reasoning. We also find striking scaling behavior: while the first hop clearly improves substantially with parameter count, the second hop (and the round-trip performance) remains relatively constant. This might indicate a fundamental limitation in today's architecture or pretraining. 

Our contributions can be summarized as follows:
\begin{itemize}[wide, labelindent=0pt, topsep=2pt, itemsep=0pt]
  \item We establish a \textbf{framework} for the investigation of \textit{latent multi-hop reasoning in LLMs} and show its \textbf{existential evidence}.
  \item We construct the \datasetname{} \textbf{dataset} which consists of 45,595 two/one-hop prompts of 52 fact composition types, created using various types of entities and relations and diverse templates~(\S\ref{sec:dataset}).
  \item We propose two novel \textbf{metrics}, \textit{internal entity recall score} and \textit{consistency score}, as proxies of the degree of the LLM's recall of an entity for its descriptive mention~(\S\ref{sec:rq1-metric}) and the degree of the LLM's utilization of its knowledge about the bridge entity's attribute~(\S\ref{sec:rq2}), respectively.
  \item We propose a \textbf{mechanism} to investigate a latent reasoning pathway even when it is not the most salient pathway determining the prediction, by measuring the relative frequency of the expected causal effects~(\S\ref{sec:rq2-setup}).
\end{itemize}

\begin{table*}[t!]
\setlength{\belowcaptionskip}{-8px}
\setlength{\tabcolsep}{3px}
\centering
\resizebox{0.9\textwidth}{!}{
    \begin{tabular}{lll}
    \toprule
    Notation & Example & Description \\
    \midrule
    $(\entx, \relx, \enty)$ & (Superstition, singer, \hhlight{Stevie Wonder}) & fact triplets of named entities where $\ei$ are named entities and $\ri$ is a\\
    $(\enty, \rely, \entz)$ & (\hhlight{Stevie Wonder}, mother, Lula) & relation function that maps $\ei$ uniquely to $\eii$, such that $\ri(\ei) = \eii$\\
    \enty{} & \hhlight{Stevie Wonder} & bridge entity that connects the two fact triplets \\
    \mpromptshort{} & ``The mother of \hhlight{Stevie Wonder} is named'' & one-hop prompt (requires one-hop reasoning) \\
    \cpromptshort{} & ``The mother of \hlight{the singer of `Superstition'} is named'' & two-hop prompt (requires two-hop reasoning) \\
    $\mt$ & ``\hlight{the singer of `Superstition'}'' & descriptive mention of the bridge entity \enty{} created with \entx{} and \relx{} \\
    - & \multirow{1}{*}{\catstr{mother of song's singer}} & \multirow{1}{*}{fact composition type} \\
    \bottomrule
    \end{tabular}
}
\caption{Notations with corresponding examples from the dataset. The text in \hhlight{brown} is the bridge entity \enty{}, Stevie Wonder (or the name of the bridge entity when presented as a substring in double quotation marks), and the text in \hlight{purple} is a descriptive mention of the bridge entity, \mt{}, ``the singer of `Superstition'''.}
\label{tab:notations-examples}
\end{table*}

\section{Related Works}\label{sec:related-works}

Recent works have shown that LLMs demonstrate remarkable in-context reasoning ability via prompting, which scales with model size~\citep{Brown2020-xt, wei2022emergent, Wei2022-qb, Zhou2022-ty}. On the contrary, when the information to reason from is not explicitly given as part of the input, LLMs often fail to correctly perform multi-hop reasoning even when they know the answer to the single-hop sub-step~\citep{Ofir-Press2023-dm,Dziri2023-vs}. While there have been wide investigations on how in-context reasoning works~\citep{chan2022data, Akyurek2022-to, Dai2022-lj, von2023transformers, Prystawski2023-ij, Feng2023-pg}, such an investigation has not been actively done to understand how latent multi-hop reasoning works.

While there have been works to investigate latent reasoning of LLMs, the exploration has been mostly done with simple single-hop reasoning tasks~\citep{Meng2022-br, Geva2023-lq, chanin2023identifying, Hernandez2023-gt} and/or controlled lightweight training/finetuning~\citep{Zhu2023-nb, Allen-Zhu2023-zc, Saparov2023-co, berglund2023taken, Berglund_undated-xe}. Also, many of the works that aim to identify latent reasoning pathways or circuits, have focused on finding the most salient reasoning pathway for simple synthetic tasks and/or toy models~\citep{Nanda2022-sz, Olsson2022-qg, brinkmann2023mechanistic, Wang2023-mz, Conmy2023-ix, Hou2023-yl, lieberum2023does, McGrath2023-co}. On the other hand, we study the existence of a latent multi-hop reasoning pathway, which may not be the most salient, in pretrained LLMs without further training, using diverse types of natural two-hop prompts.

Model editing examines ways to amend factual knowledge in LMs~\citep{De_Cao2021-ne, Mitchell2022-qn, Meng2022-br, Zhang2024-gp}. However, recent works have shown that the existing editing approaches, largely focusing on single fact edits, fail to propagate the edits to facts that depend on the edited fact~\citep{Onoe2023-wm,Zhong_undated-dp,Cohen2023-iq}. Our work explores the possibilities that such propagation could work. 
Moreover, our work investigates a pathway that affects the consistency at inference, whereas prior work in consistency has focused on quantifying inconsistency and improving consistency post-hoc~\cite{ribeiro-etal-2019-red, Li2019ALF, asai-hajishirzi-2020-logic, elazar-etal-2021-measuring, kassner-etal-2021-beliefbank, kassner-etal-2023-language, jang2023know}. \citet{Sakarvadia_2023} aim to improve multi-hop reasoning accuracy with a hypothesis that the errors stem from failure to recall the latent hop, while we investigate the foundations of this hypothesis of whether the model actually performs such a latent multi-hop reasoning. \citet{li2024understanding} is a concurrent work showing that a large portion of multi-hop reasoning failure cases can be attributed to incorrectly performing or utilizing the first hop of the latent multi-hop reasoning.

\section{Problem Formulation}

\subsection{Preliminaries}\label{sec:prelim}
We consider facts, such as \textit{``The mother of Stevie Wonder is Lula''}, as triplets $(e, r, e')$ of a subject entity $e$ (e.g., Superstition), a relation $r$ (e.g., mother), and an object entity $e'$ (e.g., Lula). Specifically, in our analysis, we focus on triplets where $e'$ is the only or the most well-known object entity for the relation $r$ for $e$ (e.g. the only mother of Stevie Wonder is Lula), and view $r$ as a function $e' = \out{r}{e}$, where \out{r}{e} is the function expression and $e'$ is the value of the expression. We analyze how LLMs process the composition of two facts with a bridge entity $\enty$ connecting them, $((\entx, \relx, \enty), (\enty, \rely, \entz))$, of which the composition is represented as \outxy{}. An example is shown in Table~\ref{tab:notations-examples}.

To query LLMs, we use a template \textify{\cdot} to convert expressions \outy{} or \outxy{} into a prompt that can be completed correctly by the value of the given expression.
For instance, the single-hop expression \out{\texttt{mother}}{\text{Stevie Wonder}} could be converted by $\textify{\out{\texttt{mother}}{\text{Stevie Wonder}}}$ to the prompt \textit{``The mother of Stevie Wonder is''}, which can be correctly completed with ``Lula''.
Similarly, the two-hop expression \out{\texttt{mother}}{\out{\texttt{singer}}{\text{Superstition}}} could be phrased by \textify{\out{\texttt{mother}}{\out{\texttt{singer}}{\text{Superstition}}}} as \textit{``The mother of the singer of `Superstition' is''} with the same correct completion.
While \mprompt{} and \cprompt{} have the same answer (``Lula''), the latter requires recalling two facts rather than one.
Therefore, we call \mprompt{} a \textit{one-hop prompt} and \cprompt{} a \textit{two-hop prompt}, and denote them as \mpromptshort{} and \cpromptshort{}, respectively.

We assume that the two-hop prompts yielded by $\textify{\cdot}$  for \outxy{} always contain a noun phrase description of the bridge entity \enty{} using \entx{} and \relx{}, e.g., \textit{``the singer of `Superstition'''} for Stevie Wonder. We denote this description as \mt{} and call it the \textit{descriptive mention} of the bridge entity \enty{}.

Last, we denote the \textit{type of the fact composition} of a two-hop prompt as {\ftype{}}, where \btype{} represents the type of the bridge entity's descriptive mention in the prompt. For example, the fact composition type of \textify{\out{\texttt{mother}}{\out{\texttt{singer}}{\text{Superstition}}}} would be \catstr{mother of song's singer}.

\subsection{Latent Multi-Hop Reasoning in LLMs}\label{sec:reasoning}

Humans possess the deductive reasoning ability to infer conclusions from given premises, such as deducing that $\outxy{} = \entz$ given a premise stating that $\outx = \enty$ and another premise stating that $\outy = \entz$. This multi-hop reasoning~\citep{welbl2018constructing, Yang2018-rb} involves identifying the bridge entity (e.g., that ``the singer of `Superstition''' is Stevie Wonder) and using it to solve for the final answer (e.g., that Stevie Wonder's mother is Lula).\looseness-1

Our research explores the extent to which a pretrained Transformer-based Large Language Model (LLM) can perform similar multi-hop reasoning when completing a two-hop prompt. Given the complex nature of LLMs, which function through high-dimensional and distributed representations, it's unlikely for a single deterministic algorithm to govern their predictions except for under highly controlled and constrained setup~\citep{Nanda2022-sz,Wang2023-mz}. Instead, LLMs may use aggregations from multiple inference pathways, ranging from shallow $n$-gram co-occurrence-based matching to deeper rule-based reasoning or even multi-hop reasoning, to make a prediction.

Therefore, to identify a pathway indicative of latent multi-hop reasoning, we focus on the internal dynamics of LLMs in processing two-hop prompts rather than the most salient pathway that contributes the most to the output.
This involves analyzing how the LLM's recall and utilization of the knowledge \outx{} and \outy{} changes in response to certain alterations made while the LLM is processing a two-hop prompt, in what we consider as the first and second hop of reasoning, respectively.

Specifically, we investigate the following two key research questions (RQs):

\begin{enumerate}[wide, labelwidth=0pt, label={\textbf{RQ\arabic*.}}, labelindent=0pt, topsep=2pt, itemsep=2pt]
    \item \textbf{How often does an LLM perform the first hop of reasoning while processing two-hop prompts?} We view the first-hop reasoning as the LLM's recall of the bridge entity for its descriptive mention. Therefore, we examine the frequency with which the LLM's internal recall of the bridge entity increases when it encounters a descriptive mention of the bridge entity within a prompt. For instance, we investigate whether altering the prompt from ``The mother of the singer of \textit{'Thriller'} is'' to {``The mother of the singer of \textit{'Superstition'} is''} increases the LLM's internal recall of Stevie Wonder.
    \item \textbf{How often does an LLM perform the second hop of reasoning while processing two-hop prompts?} We view the second-hop reasoning as the LLM's utilization of the first-hop reasoning for the second hop. Therefore, we examine the frequency with which enhancing the LLM's recall of the bridge entity for its descriptive mention improves its use of the knowledge about the bridge entity to answer the two-hop prompt. For example, we investigate if increasing the internal recall of Stevie Wonder for \textit{``the singer of `Superstition'\textquotedblright} makes the LLM better utilize its knowledge of Stevie Wonder's mother to complete the prompt.
\end{enumerate}

By addressing these questions, we aim to identify evidence of LLMs leveraging a latent pathway for multi-hop reasoning.
\section{\datasetname{} Dataset}\label{sec:dataset}
To answer our questions with prompts of diverse fact composition types, we construct \datasetname{} using well-known named entities in Wikidata~\citep{vrandevcic2014wikidata} and manually selected relations (Appendix~\ref{sec:dataset-construction}). \datasetname{} consists of 45,595 unique pairs of one-hop and two-hop prompts of 52 fact composition types constructed from the same number of fact triplet pairs $((\entx, \relx, \enty), (\enty, \rely, \entz))$ as in Table~\ref{tab:notations-examples}. Appendix Table~\ref{tab:dataset} shows example two-hop prompts for each fact composition type, and Appendix~\ref{sec:dataset-stats} provides detailed data statistics.

\section{First Hop of Multi-Hop Reasoning}\label{sec:rq1}
In this section, we answer RQ1 of \textit{how often an LLM performs the first hop of reasoning while processing two-hop prompts}. We first introduce \xmetricop{} as a metric to approximate the LLM's internal recall of the bridge entity 
upon its descriptive mention in a prompt~(\S\ref{sec:rq1-metric}). Next, we propose to measure how often this recall increases when changing the input prompt to indirectly mention the bridge entity~(\S\ref{sec:rq1-setup}). Then, we evaluate this using \datasetname{} and answer RQ1~(\S\ref{sec:rq1-results}).

\subsection{Internal Entity Recall Score}\label{sec:rq1-metric}
We define \xmetricop{} as a metric to measure the LLM's recall of the bridge entity \enty{} within a two-hop prompt \cpromptshort{}. This is defined with respect to the hidden representation in a certain layer $l$, at the last position of the bridge entity's descriptive mention in the two-hop prompt. This hidden representation is projected to the vocabulary space to calculate the log probability of the first token of the entity's name (e.g., the first token of ``Stevie Wonder'').
Formally, let \mentyfirst{} be the first token of \enty{}, then:
\begin{align}\label{eq:xmetricreal}
    &\xmetricreal{l} \\
    &= \log \softmax(\layernorm(\residstream{l}) W_U)_{\idx(\mentyfirst{})},\nonumber
\end{align}
\noindent
where $\residstream{l}\in \dims{h}$ is the output from the $l$-th Transformer layer at the last token of the bridge entity's descriptive mention in the two-hop prompt $\cpromptshort$, 
and $\idx(\mentyfirst{}) \in [0, V-1]$ is the index of the token \mentyfirst{} in the unembedding matrix $W_U \in \dims{h \times V}$. $\layernorm{}$ is the layer normalization used for the last layer output \residstream{L-1} before projecting it to the unembedding matrix to obtain the output next-token probability distribution. Applying this normalization makes \xmetricreal{L-1} compatible with 
the output probability of \mentyfirst{} as the next token of the prefix of \cpromptshort{} ending at the descriptive mention (e.g., ``The mother of the singer of `Superstition''').\footnote{We omit the bias term as it often models the frequency of the token~\citep{kobayashi-etal-2023-transformer}, which we do not want to consider for measuring the internal recall of an entity.}
We interpret higher \xmetricreal{l} as stronger internal recall of the bridge entity \enty{} at the $l$-th layer.

The proposed definition of \xmetricop{} is inspired by previous works which report that the representation constructed at the last token position of a subject often plays an important role in encoding information about the subject~\citep{Meng2022-br, Geva2023-lq}, the work of \citet{nostalgebraist2020} that projects early-layer outputs to the vocabulary space, and the work of \citet{Geva2022-ah} which shows that such projections at the last subject token position of one-hop prompts provide interpretable top-rank attributes that are semantically relevant to the subject. Although \xmetricop{} assesses the recall of an entity with respect to only the first token of its name, 
it is directly related to how auto-regressive LLMs process the input text and prepare the next token to generate. A control experiment in Appendix~\ref{sec:apposition} validates \xmetricop{} as a reasonable proxy for measuring the internal entity recall.

\subsection{Experiment}\label{sec:rq1-setup}

Given \xmetricop{}, we answer RQ1 by measuring how often the internal recall of \enty{} improves at layer $l$ when modifying a two-hop prompt from \negcprompt{} to \cpromptshort{}, where \negcprompt{} does not contain the descriptive mention of \enty{} while \cpromptshort{} does. To be specific, we measure \textit{the relative frequency} of \cpromptshort{} in \datasetname{} where $\xmetricop^l(\enty, \cpromptshort) > \xmetricop^l(\enty,\negcprompt).$

To construct \negcprompt{}, we alter the descriptive mention of the bridge entity in \cpromptshort{} in two ways: by replacing \entx{} with \negentx{} such that $\mentionize{\out{\relx}{\negentx}}$ does not point to \enty{}, or \relx{} with \negrelx{} to ensure $\mentionize{\out{\negrelx}{\entx}}$ does not refer to \enty{}. Examples include substituting ``the singer of `Superstition''' in \cpromptshort{} to ``the singer of \textit{`Thriller'}'' or ``\textit{a plagiarist} of `Superstition'''. These adjustments are termed \textit{entity substitution} and \textit{relation substitution}, respectively.

For each two-hop prompt \cpromptshort{} in \datasetname{}, we randomly select one \negentx{} from the same fact composition type and one \negrelx{} from a set of predefined candidate relations (provided in Appendix Table~\ref{tab:negrelx}) to create \negcprompt{}. We then measure the relative frequency of cases where replacing \negcprompt{} with \cpromptshort{} via entity or relation substitution increases the recall of \enty{}. A relative frequency above 0.5 suggests the LLM's chance to perform first-hop reasoning exceeds the random chance for these prompts.

\subsection{Results}\label{sec:rq1-results}
\begin{figure*}[t!]
\centering
    \begin{subfigure}[t]{0.24\textwidth}
    \includegraphics[width=\textwidth]{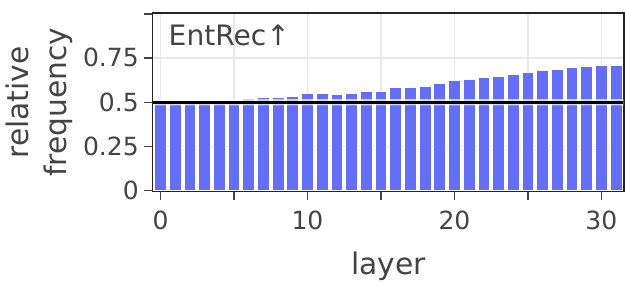}
    \caption{7B entity substitution}\label{fig:negentx}
    \end{subfigure}\hfill
    \begin{subfigure}[t]{0.24\textwidth}
    \includegraphics[width=\textwidth]{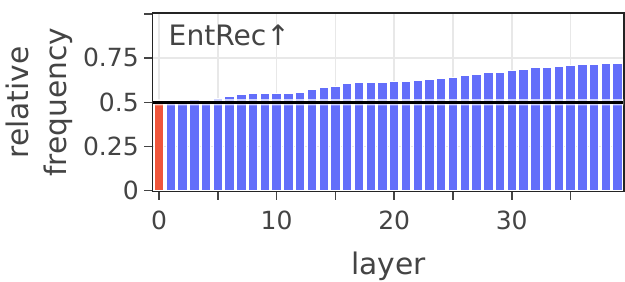}
    \caption{13B entity substitution}\label{fig:negentx_13b}
    \end{subfigure}\hfill
    \begin{subfigure}[t]{0.24\textwidth}
    \includegraphics[width=\textwidth]{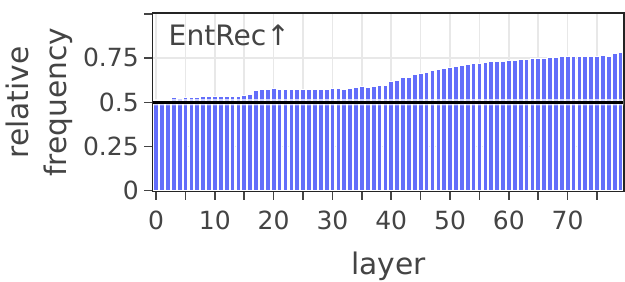}
    \caption{70B entity substitution}\label{fig:negentx_70b}
    \end{subfigure}\hfill
    \begin{subfigure}[t]{0.24\textwidth}
    \includegraphics[width=\textwidth]{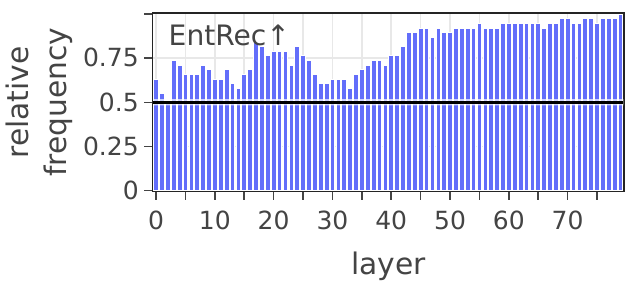}
    \caption{70B entity substitution for \catstr{president of anthem's country}}\label{fig:negent_pres}
    \end{subfigure}\hfill
    \setlength{\belowcaptionskip}{-8px}
    \begin{subfigure}[t]{0.24\textwidth}
    \includegraphics[width=\textwidth]{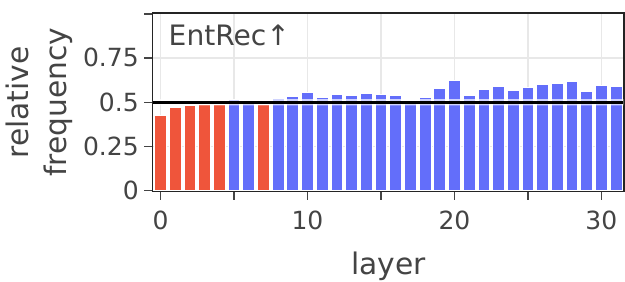}
    \caption{7B relation substitution}\label{fig:negrelx}
    \end{subfigure}\hfill
    \begin{subfigure}[t]{0.24\textwidth}
    \includegraphics[width=\textwidth]{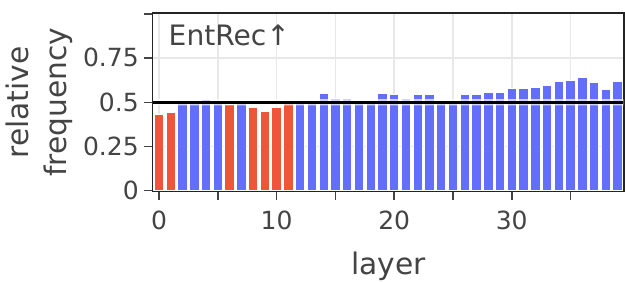}
    \caption{13B relation substitution}\label{fig:negrelx_13b}
    \end{subfigure}\hfill
    \begin{subfigure}[t]{0.24\textwidth}
    \includegraphics[width=\textwidth]{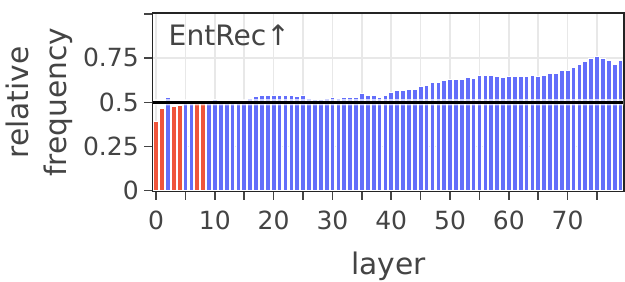}
    \caption{70B relation substitution}\label{fig:negrelx_70b}
    \end{subfigure}\hfill
    \begin{subfigure}[t]{0.24\textwidth}
    \includegraphics[width=\textwidth]{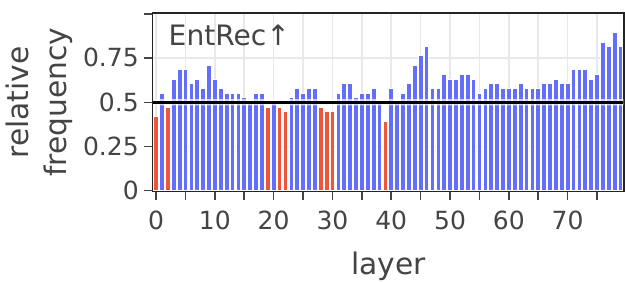}
    \caption{70B relation subst. for \catstr{president of anthem's country}}\label{fig:negrel_pres}
    \end{subfigure}
\caption{Relative frequency of the cases where the internal recall of the bridge entity of LLaMA-2 increases with entity substitution (top row) and relation substitution (bottom row). Bars are colored blue if the relative frequency is greater than or equal to 0.5 and red otherwise.}
\label{fig:rq1_main_ent}
\end{figure*}
\begin{figure*}[t!]
\setlength{\belowcaptionskip}{-8px}
\centering
    \begin{subfigure}[t]{0.195\textwidth}
    \includegraphics[width=\textwidth]{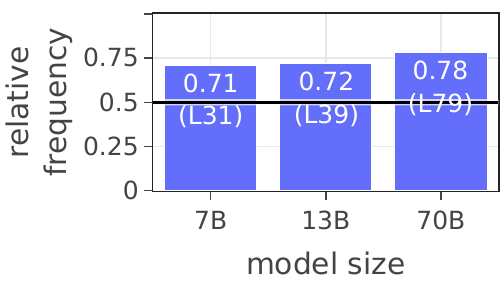}
    \caption{RQ1 entity substitution result (\S\ref{sec:rq1})}\label{fig:result_rq1_ent}
    \end{subfigure}
    \hfill
    \begin{subfigure}[t]{0.195\textwidth}
    \includegraphics[width=\textwidth]{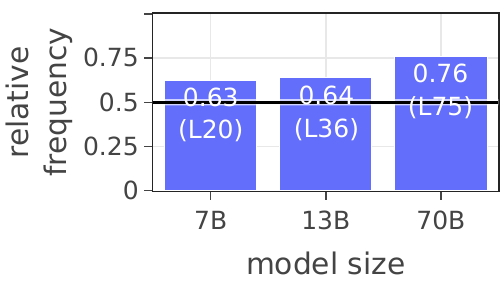}
    \caption{RQ1 relation substitution result (\S\ref{sec:rq1})}\label{fig:result_rq1_rel}
    \end{subfigure}
    \hfill
    \begin{subfigure}[t]{0.195\textwidth}
    \includegraphics[width=\textwidth]{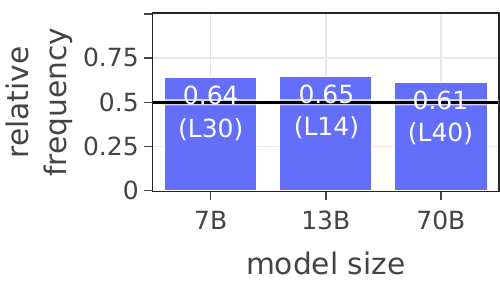}
    \caption{RQ2 result (\S\ref{sec:rq2})}\label{fig:result_rq2}
    \end{subfigure}
    \hfill
    \begin{subfigure}[t]{0.195\textwidth}
    \includegraphics[width=\textwidth]{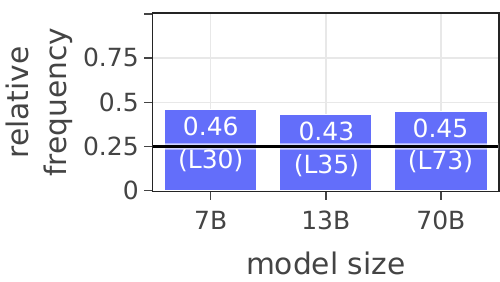}
    \caption{RQ1\&2 entity substitution result (\S\ref{sec:rq12})}\label{fig:result_rq12_ent}
    \end{subfigure}
    \hfill
    \begin{subfigure}[t]{0.195\textwidth}
    \includegraphics[width=\textwidth]{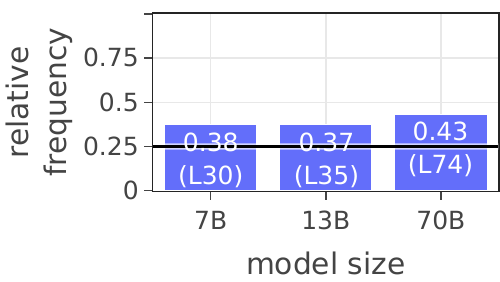}
    \caption{RQ1\&2 relation substitution result~(\S\ref{sec:rq12})}\label{fig:result_rq12_rel}
    \end{subfigure}
\caption{Experimental results with increasing scale of LLaMA-2. Technical details for all experiments in our work can be found in Appendix~\ref{sec:technical-details}.}
\label{fig:result}
\end{figure*}

\paragraph{There is substantial evidence of the first hop of reasoning, which becomes stronger with increasing model size.} Figure~\ref{fig:rq1_main_ent} shows the relative frequency of the cases that the entity recall at each layer increases with entity and relation substitution. LLaMA-2 7B entity substitution result~(Figure~\ref{fig:negentx}) shows that the evidence of first-hop reasoning becomes clearer with increasing layer depth, peaking at 0.71 in layer 31. Relation substitution exhibits a slightly noisier pattern with a peak at 0.63 in layer 20 (Figure~\ref{fig:negrelx}).

As model size increases from 7B to 13B and 70B, first-hop reasoning occurs more frequently for both entity substitution and relation substitution. For the former, the maximum relative frequency rises from 0.71 (7B) to 0.72 (13B) and 0.78 (70B)~(Figure~\ref{fig:result_rq1_ent}). For the latter, it increases from 0.63 (7B) to 0.64 (13B) and 0.76 (70B)~(Figure~\ref{fig:result_rq1_rel}).

\paragraph{Relatively strong evidence supports the first-hop reasoning in up to 73\% of fact composition types.}
With LLaMA-2 7B-13B-70B, 18/25/34 and 21/27/38 out of 52 of fact composition types exhibit maximum relative frequencies exceeding 0.8 for entity and relation substitution, respectively. In addition, 11 out of 52 types demonstrate such strong first-hop reasoning evidence robustly across all model sizes and substitution types. For example, the maximum frequency of \catstr{president of anthem's country} (``The country with the national anthem `Azat u ankakh Artsakh' is led by president'') shows the maximum frequency of 0.97/0.92/1.0~(Figure~\ref{fig:negent_pres}) and 0.87/0.87/0.89~(Figure~\ref{fig:negrel_pres}) with each model and substitution, respectively. Individual fact composition types exhibit diverse patterns of relative frequency across layers.
\vspace{-0.2em}
\section{Second Hop of Multi-Hop Reasoning}\label{sec:rq2}
\vspace{-0.1em}
In this section, we answer RQ2 of \textit{how often an LLM performs the second-hop reasoning while processing two-hop prompts}.
We view the second hop of reasoning as the LLM's utilization of what it knows about the bridge entity's attribute (Stevie Wonder's mother) to answer the two-hop prompt about the same attribute of the entity referred to by the descriptive mention (the singer of `Superstition''s mother).
Therefore, when an LLM performs the second hop, we expect to see a connection between its recall of the bridge entity (i.e. resolving the first hop) and its similarity in responding to a two-hop prompt and a corresponding one-hop prompt about the bridge entity's attribute, e.g., the two-hop prompt \textit{``The mother of the singer of `Superstition' is''} and the one-hop prompt \textit{``The mother of Stevie Wonder is''}.
Namely, the more strongly the model recalls the bridge entity (e.g., Stevie Wonder) while processing the two-hop prompt, the more similar the completion of this prompt should be to the completion of the one-hop prompt. In the following, we describe our approach for testing how often such a causal connection exists between entity recall and the \textit{similarity} in the prompt completions, which we refer to as \textit{consistency}.\looseness-1
\vspace{-0.2em}
\subsection{Consistency Score}\label{sec:rq2-metric}

We define \ymetricop{} to measure how consistently an LLM responds to the two-hop and one-hop prompts. Let $\mathbf{p}_{\cpromptshort}, \mathbf{p}_{\mpromptshort} \in \dims{V}$ be the output probability distributions for a two-hop prompt $\cpromptshort{}$ and the corresponding one-hop prompt $\mpromptshort{}$, respectively. Denoting $\mathrm{H}(Q, P) = -\sum_{i=0}^{V-1} P_i\log Q_i$ as the cross-entropy between probability distributions $P$ and $Q$, we define:
\begin{align}\label{eq:ymetricreal}
\begin{split}
    &\ymetricreal{} \\
    &= -0.5 \mathrm{H}(\mathbf{p}_{\cpromptshort}, \mathbf{p}_{\mpromptshort}) -0.5 \mathrm{H}(\mathbf{p}_{\mpromptshort}, \mathbf{p}_{\cpromptshort}).
\end{split}
\end{align}
This score evaluates the similarity between the two probability distributions by computing and averaging their cross-entropy, ensuring symmetry in the evaluation. The symmetry from averaging mitigates sensitivity to the individual distribution's entropy levels, aiming for equal treatment of divergences in both directions.

Note that we use consistency instead of two-hop prompt completion accuracy or the probability of the ground truth answer because the latter metrics are insufficient to capture the second-hop reasoning for the cases where the corresponding one-hop prompt completion is incorrect. In addition, these metrics inherit noise from the choice of the ground truth answer or the set of answer candidates. On the other hand, comparing the similarity of the output distributions is not affected by the choice of ground truth, and provides a way to capture the second-hop reasoning even when the ground truth answer is not in the top-1 generation of the one-hop prompt.

Also, we do not choose to compare the completion strings or their binary accuracy of the one/two-hop prompts because these metrics cannot capture subtle consistency differences in the probability distribution. We choose cross-entropy rather than Kullback–Leibler or Jensen-Shannon divergence because the latter metrics contain an entropy term that is irrelevant to consistency, but can dominate the score, diluting the cross-entropy signal. Higher consistency scores indicate greater similarity between the output distributions. In Appendix~\ref{sec:cot}, we provide empirical evidence for the consistency score being a reasonable approximation of the utilization of the model's knowledge about the bridge entity's attribute.
\begin{figure*}[t!]
\centering
    \begin{subfigure}[t]{0.24\textwidth}
    \includegraphics[width=\textwidth]{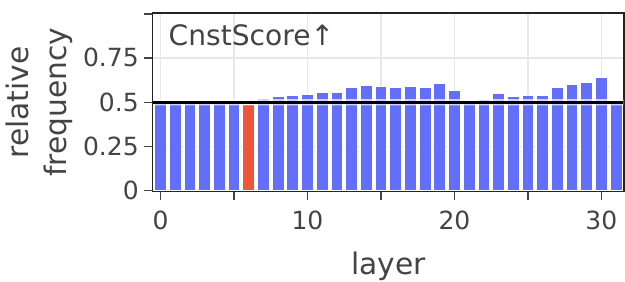}
    \caption{LLaMA-2 7B}\label{fig:rq2_consistency}
    \end{subfigure}
    \hfill
    \begin{subfigure}[t]{0.24\textwidth}
    \includegraphics[width=\textwidth]{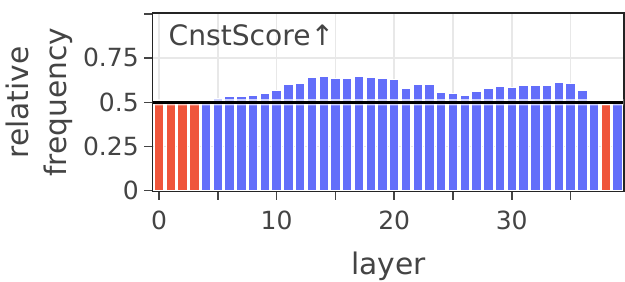}
    \caption{LLaMA-2 13B}\label{fig:rq2_consistency_13b}
    \end{subfigure}
    \hfill
    \begin{subfigure}[t]{0.24\textwidth}
    \includegraphics[width=\textwidth]{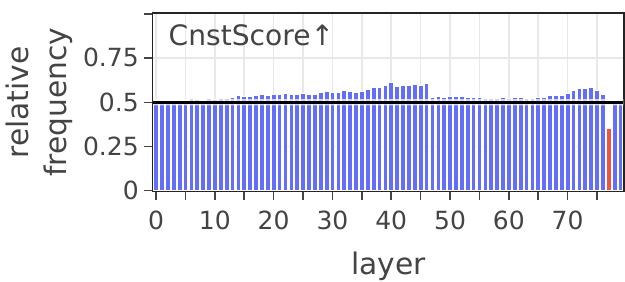}
    \caption{LLaMA-2 70B}\label{fig:rq2_consistency_70b}
    \end{subfigure}
    \hfill
    \begin{subfigure}[t]{0.24\textwidth}
    \includegraphics[width=\textwidth]{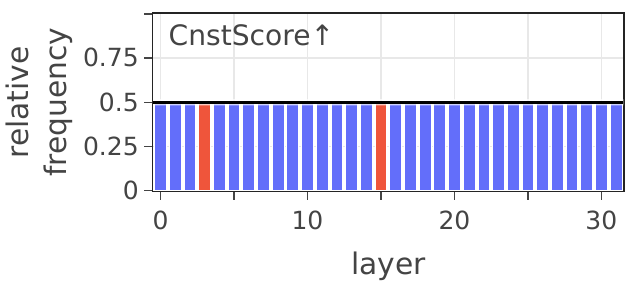}
    \caption{Random LLaMA-2 7B}\label{fig:rq2_consistency_scratch}
    \end{subfigure}\hfill
    \setlength{\belowcaptionskip}{-8px}
    \begin{subfigure}[t]{0.24\textwidth}
    \includegraphics[width=\textwidth]{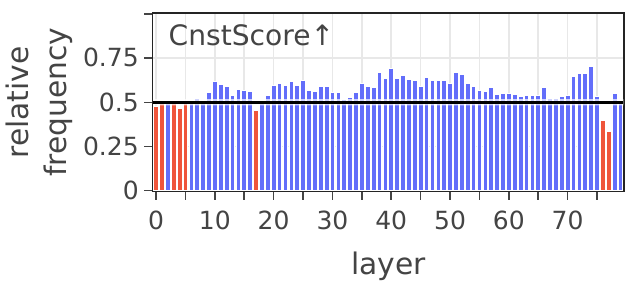}
    \caption{70B result of \catstr{stock exchange of game's developer}}
    \end{subfigure}\hfill
    \begin{subfigure}[t]{0.24\textwidth}
    \includegraphics[width=\textwidth]{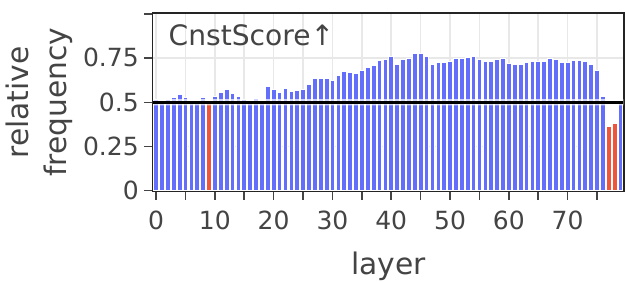}
    \caption{70B result of \catstr{mother of song's singer}}
    \end{subfigure}\hfill
    \begin{subfigure}[t]{0.24\textwidth}
    \includegraphics[width=\textwidth]{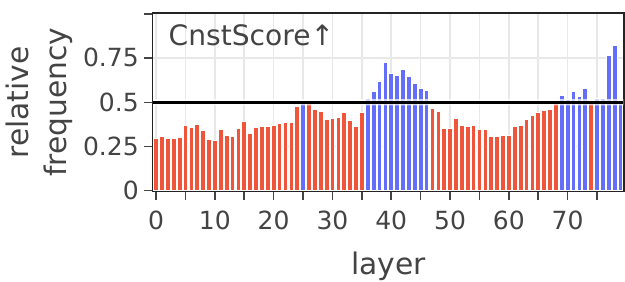}
    \caption{70B result of \catstr{founder of person's undergrad university}}\label{fig:rq2_founder}
    \end{subfigure}\hfill
    \begin{subfigure}[t]{0.24\textwidth}
    \includegraphics[width=\textwidth]{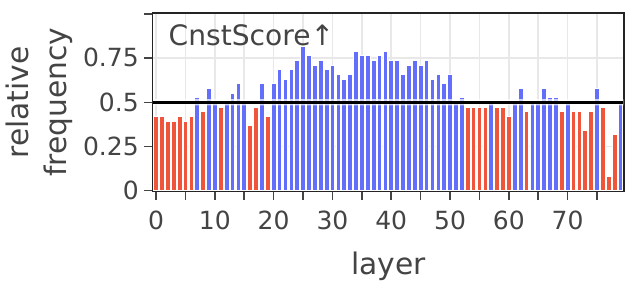}
    \caption{70B result of \catstr{president of anthem's country}}\label{fig:rq2_president}
    \end{subfigure}\hfill
\caption{Relative frequency that stronger recall of the bridge entity at the $l$-th layer increases the consistency of the LLM. Bars are colored blue if the relative frequency is greater than or equal to 0.5 and red otherwise. We manually set the value of 0.5 at the last layer because the intervention does not affect the consistency at that layer.}
\label{fig:rq2_main}
\end{figure*}

\subsection{Experiment}\label{sec:rq2-setup}
Given \xmetricop{} and \ymetricop{}, we answer RQ2 by measuring how often increasing the recall of the bridge entity \enty{} at the $l$-th layer increases the LLM's consistency in answering the two-hop prompt with respect to the one-hop prompt. In other words, we examine whether increasing \xmetricreal{l} leads to increasing \ymetricreal{}.

We would have been able to use differential calculus to obtain the answer by calculating the direction of change if \ymetricreal{} were directly dependent on \xmetricreal{l}. However, there exists no direct functional dependency between the two values. Instead, we leverage the shared reliance of both metrics on \residstream{l} for computation where $l \in [0, L-1)$,\footnote{\ymetricreal{} utilizes $\mathbf{p}_{\cpromptshort}$, which utilizes \residstream{l} for its calculation. However, only $\residstream{l} \text{ where } l = 0, \cdots, L-2$ are used to calculate the attention outputs at layers $l = 1, \cdots, L-1$, respectively, to get $\mathbf{p}_{\cpromptshort}$.} redefining them as \xmetricreparam{} and \ymetricreparam{} relative to \residstream{l}. This reparameterization allows us to change the question to: if \xmetricreparam{} is increased by altering \residstream{l}, does \ymetricreparam{} also increase?

To explore this, we adjust \xmetricreparam{} in the direction of its steepest increase, represented by $\nabla_{\residstream{l}} \xmetricreparam{}$, and observe the impact on \ymetricreparam{} by modifying \residstream{l} according to a magnitude of change $\alpha$:
\[
\newresidstream{l}(\alpha) = \residstream{l} + \alpha \nabla_{\residstream{l}} \xmetricreparam{}.
\]
Subsequently, we calculate \ymetricreparam{} using $\newresidstream{l}(\alpha)$,\footnote{We use activation patching~\citep{Wang2023-mz} to implement the replacement of \residstream{l} with $\newresidstream{l}(\alpha)$.} which allows us to express it as a function \ymetricfunc{\alpha} of $\alpha$. Then, we examine its derivative, $\left. \frac{d}{d\alpha}\ymetricfunc{\alpha} \right|_{\alpha=0}$ to understand the direction of change at the current value. A positive derivative indicates that an increase in \xmetricreparam{} leads to an increase in \ymetricreal{}, while a negative one suggests the opposite. By assessing \textit{the relative frequency of positive gradients} among the two-hop prompts in \datasetname{}, we quantify how often the LLM performs the second hop of the reasoning, with frequencies above 0.5 suggesting that the LLM's chance to perform the second-hop reasoning exceeds random chance for these prompts.

\subsection{Results}\label{sec:rq2-results}

\paragraph{There is moderate evidence of the second-hop reasoning, which does not become stronger with increasing model size.}
Figure~\ref{fig:rq2_main} shows the relative frequency of the cases where increasing the bridge entity recall increases the consistency. In LLaMA-2 7B, the middle and late layers exhibit a relative frequency higher than 0.5 (random chance) with statistical significance, peaking at 0.64 in layer 30. Test result with a randomly initialized model verifies 0.5 as the randomness baseline~(Figure~\ref{fig:rq2_consistency_scratch}).

However, unlike the first-hop reasoning (\S\ref{sec:rq1}), the second-hop reasoning does not strengthen with increasing model size; when scaling from 7B to 13B and 70B, the maximum relative frequency remains relatively stable at 0.64 (7B), 0.65 (13B), and 0.61 (70B), as shown in Figure~\ref{fig:result_rq2}. This observation does not change even when the test is conducted using the log probability of the ground truth answer instead of \ymetricop{} (Appendix~\ref{sec:accuracy}). It is worth noting that this finding aligns with the observation of \citet{Ofir-Press2023-dm}, that the single-hop question answering performance improves faster than the multi-hop performance as the model size increases, and thus the \textit{compositionality gap} (the ratio of how often models can correctly answer all sub-problems but not generate the overall solution) does not decrease with increasing model size.

\paragraph{Relatively strong evidence supports the second-hop reasoning in up to 19\% of fact composition types.}
With LLaMA-2 7B-13B-70B, 10/7/5 out of 52 of fact composition types exhibit maximum relative frequencies exceeding 0.8, respectively. Among them, \catstr{founder of person's undergraduate university} and \catstr{president of anthem's country} demonstrate such strong second-hop reasoning evidence across all model sizes, with a maximum frequency of 0.86/0.81/0.82 (Figure~\ref{fig:rq2_founder}) and 0.84/0.89/0.82 (Figure~\ref{fig:rq2_president}), respectively.

\begin{figure*}[t!]
\centering
    \begin{subfigure}[t]{0.85\textwidth}
    \includegraphics[width=\textwidth]{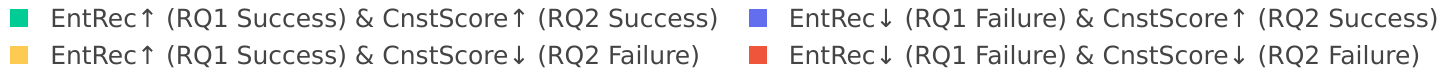}
    \end{subfigure}\\
    \begin{subfigure}[t]{0.24\textwidth}
    \includegraphics[width=\textwidth]{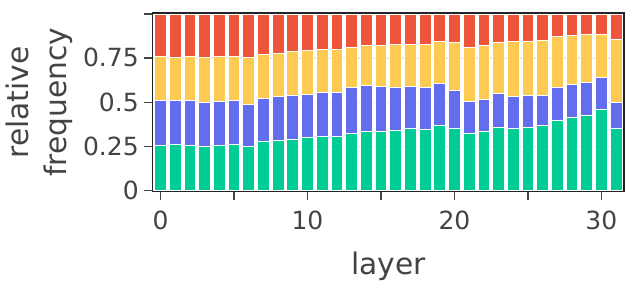}
    \caption{7B entity substitution}\label{fig:rq_negent}
    \end{subfigure}
    \hfill
    \begin{subfigure}[t]{0.24\textwidth}
    \includegraphics[width=\textwidth]{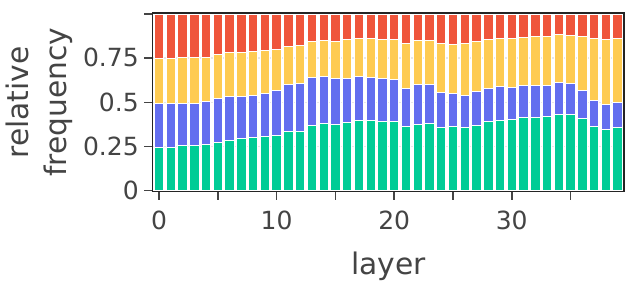}
    \caption{13B entity substitution}\label{fig:rq_negent_13b}
    \end{subfigure}
    \hfill
    \begin{subfigure}[t]{0.24\textwidth}
    \includegraphics[width=\textwidth]{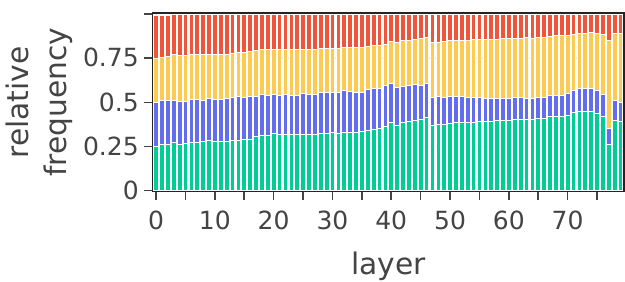}
    \caption{70B entity substitution}\label{fig:rq_negent_70b}
    \end{subfigure}
    \hfill
    \begin{subfigure}[t]{0.24\textwidth}
    \includegraphics[width=\textwidth]{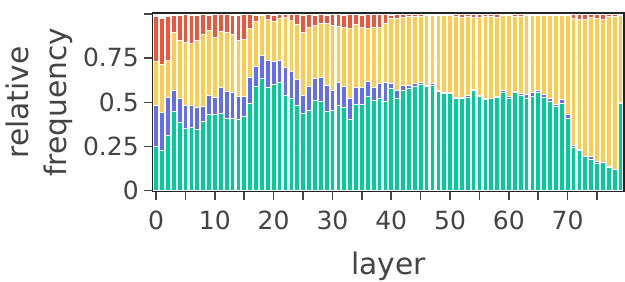}
    \caption{70B entity substitution for \catstr{anthem of capital's country}}\label{fig:rq_ent_pres}
    \end{subfigure}\hfill
    \setlength{\belowcaptionskip}{-8px}
    \begin{subfigure}[t]{0.24\textwidth}
    \includegraphics[width=\textwidth]{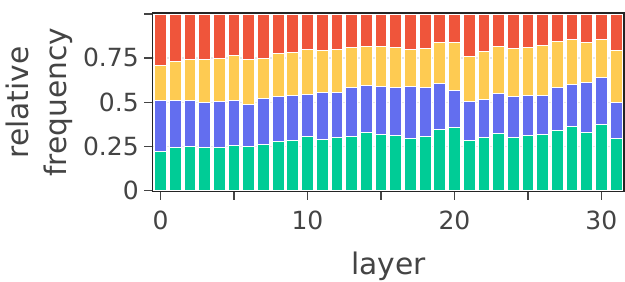}
    \caption{7B relation substitution}\label{fig:rq_negrel}
    \end{subfigure}
    \hfill
    \begin{subfigure}[t]{0.24\textwidth}
    \includegraphics[width=\textwidth]{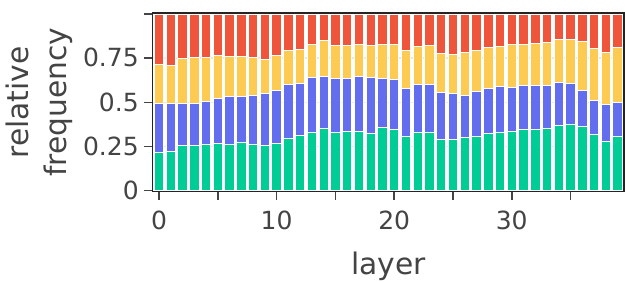}
    \caption{13B relation substitution}\label{fig:rq_negrel_13b}
    \end{subfigure}
    \begin{subfigure}[t]{0.24\textwidth}
    \includegraphics[width=\textwidth]{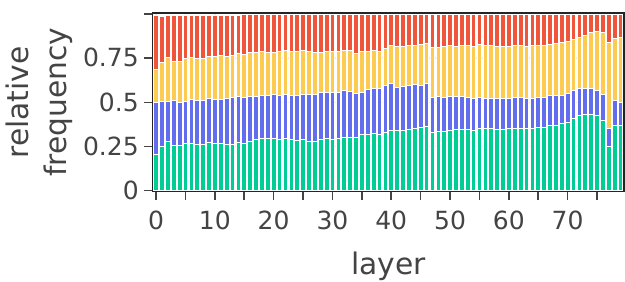}
    \caption{70B relation substitution}\label{fig:rq_negrel_70b}
    \end{subfigure}
    \hfill
    \begin{subfigure}[t]{0.24\textwidth}
    \includegraphics[width=\textwidth]{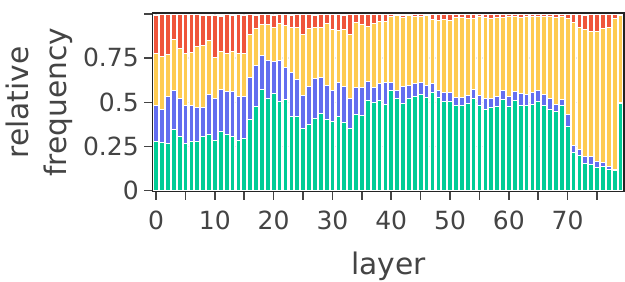}
    \caption{70B relation subst. for \catstr{anthem of capital's country}}\label{fig:rq_rel_pres}
    \end{subfigure}
\caption{Relative frequency of the four outcomes of RQ1 and RQ2 in LLaMA-2 models, with entity substitution (top row) and relation substitution (bottom row) for RQ1. Let the increase of the entity recall with the input substitution for the first hop reasoning be the \textit{success} case of RQ1, and the increase of the consistency score with the increased entity recall for the second hop reasoning be the \textit{success} case of RQ2. The green, blue, yellow, and red bars show the cases of SS (success-success), FS, SF, and FF for RQ1 and RQ2, respectively. We manually set the value of the last layer as 0.5 multiplied by the relative frequency for RQ1 because the intervention does not affect the consistency at that layer.}\label{fig:rq_cases}
\end{figure*}

\section{Latent Multi-Hop Reasoning}\label{sec:rq12}
In this section, we measure \textit{how often LLMs perform latent multi-hop reasoning while processing the two-hop prompt} by combining our answers to RQ1 and RQ2. For each two-hop prompt, we consider successful outcomes for RQ1 (an entity recall increase with entity/relation substitution) and RQ2 (a consistency increase with increased entity recall) as evidence of the first and second hops of reasoning, respectively. Four possible outcomes arise: (SS) success in both RQ1 and RQ2 that we view as the multi-hop reasoning; (FS) failure in RQ1 but success in RQ2; (SF) success in RQ1 but failure in RQ2; (FF) failure in both RQ1 and RQ2.

\paragraph{There is moderate evidence of the latent multi-hop reasoning, which sometimes becomes stronger with increasing model size.} Figure~\ref{fig:rq_cases} shows the relative frequency of the four cases, where green, blue, yellow, and red represent each of the cases of SS, FS, SF, and FF, respectively. LLaMA-2 7B exhibits a relative frequency for successful multi-hop reasoning (green) above random chance (0.25), peaking at 0.46 (entity substitution) and 0.38 (relation substitution). The likelihood of partial multi-hop reasoning (green + blue + yellow) exceeds 0.8 in later layers.

While entity substitution results do not show increased multi-hop reasoning with model size (Figure~\ref{fig:result_rq12_ent}), relation substitution exhibits a scaling trend. From 7B to 70B, the maximum relative frequency increases from 0.38 to 0.43, suggesting that larger models may facilitate multi-hop reasoning with relational changes (Figure~\ref{fig:result_rq12_rel}).

\paragraph{Relatively strong evidence supports latent multi-hop reasoning in up to 23\% of fact composition types.} Considering $0.8^2 = 0.64$ as the threshold, with respect to LLaMA-2 7B-13B-70B, 7/3/12 types exceed the threshold with entity substitution and 3/3/9 types do so with relation substitution. The maximum frequency of \catstr{anthem of capital's country} (``The national anthem of the country led by president Lazarus Chakwera is named'') exceeds this threshold across all models and substitutions with 0.68/0.82/0.66 (Figure~\ref{fig:rq_ent_pres}) and 0.74/0.82/0.68 (Figure~\ref{fig:rq_rel_pres}), respectively. Individual types show diverse patterns distinct from the overall dataset.

\section{Discussion and Conclusion}
Our work studies the latent multi-hop reasoning abilities of LLMs. We find strong evidence of latent multi-hop reasoning for certain fact composition types with the reasoning pathway utilized in more than 80\% of the cases. However, the utilization is highly contextual; there are also fact composition types where we see weak or almost no evidence of reasoning. The evidence of second and multi-hop reasoning across the whole set of prompts is rather moderate and only substantial in the first hop.

Moreover, while we see a clear scaling trend with the first hop of the latent multi-hop reasoning pathway with increasing model size, we do not see such scaling evidence for the second-hop reasoning pathway. This could be the reason behind the observation of \citet{Ofir-Press2023-dm} that the compositionality gap (the ratio of how often models can correctly answer all sub-problems but not generate the overall solution) does not decrease with increasing model size.

Although our analysis is based on LLaMA-2 family of models of up to 70B parameters, our findings suggest potential limitations in the current scaling paradigm for promoting latent multi-hop reasoning.
Thus, we may need to study the choice of pretraining data, loss functions that promote knowledge retrieval and utilization, or model architectures with a stronger inductive bias towards internal knowledge representation for LLMs' stronger latent reasoning abilities. However, analyzing the subset of prompts with strong evidence of multi-hop reasoning with respect to pretraining dynamics and data may give insights into the emergence of such abilities even in the context of the current pretraining and scaling paradigm.

Overall, our findings advance the understanding of LLM capabilities and can guide future research aiming to promote and strengthen latent multi-hop reasoning which is relevant for parameter efficiency, generalization, and controllability.

\section{Limitations}
\paragraph{Latent Multi-Hop Reasoning Pathway} While we study one pathway for latent multi-hop reasoning (e.g., we test the use of the second hop by means of entity recall), considering the potential redundancy of inference pathways in LLMs~\citep{McGrath2023-co}, other pathways might exist; the same information might be retrieved in different ways. Also, we don't measure multi-hop reasoning end-to-end and track only the changes that occur in the first and the second hop with respect to a single layer, while the effect of the first hop of reasoning could possibly propagate to other layers. Hence, the effects we see might be a lower bound on the model's ability to perform latent two-hop reasoning.

\paragraph{Dataset} We aim to collect fact triplets $(e, r, e')$ such that $e' = r(e)$ is the only or the most famous object for the relation $r$ for $e$. Although we use the entities with the most number of reference links and ensure that $e'$ is the only object entity at least among the collected fact triplets for this purpose, there are noises introduced from Wikidata. Besides, in reality, it is difficult to strictly satisfy the condition of ``only'' due to the vast amount of real-world knowledge that changes rapidly and dynamically.

\paragraph{Metrics} Our measure of internal entity recall is an approximation as we use only the first token of the entity, although it is directly related to how LLMs process the input text and prepare the next token to generate. Moreover, the internal entity recall score is based on logit lens~\citep{nostalgebraist2020} which has shortcomings such as representation drift, bias, and brittleness~\citep{tuned_lens, Timkey2021}. However, these limitations have minimal effect on our analysis because our focus is not on making the prediction accurate in early layers as studied for adaptive computation methods such as early exit~\citep{Din_Geva}, but to study the LLM's internal dynamics as-is.\looseness-1

\section*{Acknowledgments}
We would like to thank Sang-Woo Lee, Jasmijn Bastings, William Cohen, and Owain Evans for the valuable feedback and discussions.

\bibliography{custom}

\clearpage
\appendix

\begin{table*}[t!]
\centering
\resizebox{0.99\textwidth}{!}{
    \begin{tabular}{lll}
    \toprule
    Term & Notation & Example \\
    \midrule
    fact composition type & \ftype{} & \catstr{birth city of novel's author} \\
    first fact triplet & $(\entx, \relx, \enty)$ & (Ubik, author, Philip K. Dick) \\
    second fact triplet & $(\enty, \rely, \entz)$ & (Philip K. Dick, birth city, Chicago) \\
    \gcmidrule(lr){1-3}
    mention-constructing template & \templatex{\cdot} & $\templateforx_\text{author}({\cdot}) = \text{``the author of the novel $\cdots$''}$ \\   
    prompt-constructing template & \templatey{\cdot} & $\templatefory_\text{birth city}({\cdot}) = \text{``$\cdots$ was born in the city of''}$ \\\gcmidrule(lr){1-3}
    descriptive mention of \enty{} & $\mt{} = \templatex{\name{\entx}}$ & $\templateforx_\text{author}(\name{\text{Ubik}}) = \text{\hlight{``the author of the novel Ubik''}}$ \\
    two-hop prompt & $\cprompt = \templatey{\templatex{\name{\entx}}}$ & $\templatefory_\text{birth city}(\templateforx_\text{author}(\name{\text{Superstition}})) = \text{``\hlight{The author of the novel Ubik} was born in the city of''}$ \\
    one-hop prompt & $\mprompt = \templatey{\name{\enty}}$ & $\templatefory_\text{birth city}(\name{\text{Philip K. Dick}}) = \text{``\hhlight{Philip K. Dick} was born in the city of''}$ \\
    \midrule
    fact composition type & \ftype{} & \catstr{director of main character's movie} \\
    first fact triplet & $(\entx, \relx, \enty)$ & (Dominick Cobb, movie, Inception) \\
    second fact triplet & $(\enty, \rely, \entz)$ & (Inception, director, Christopher Nolan) \\\gcmidrule(lr){1-3}
    mention-constructing template & \templatex{\cdot} & $\templateforx_\text{movie}({\cdot}) = \hlight{\text{``the movie featuring $\cdots$ as the main character''}}$ \\   
    prompt-constructing template & \templatey{\cdot} & $\templatefory_\text{director}({\cdot}) = \text{``The name of the director of $\cdots$ is''}$ \\\gcmidrule(lr){1-3}
    descriptive mention of \enty{} & $\mt{} = \templatex{\name{\entx}}$ & $\templateforx_\text{movie}(\name{\text{Dominick Cobb}}) = \text{``the movie featuring Dominick Cobb as the main character''}$ \\
    two-hop prompt & $\cprompt = \templatey{\templatex{\name{\entx}}}$ & $\templatefory_\text{director}(\templateforx_\text{movie}(\name{\text{Dominick Cobb}}))$ \\
    &  & $= \text{``The name of the director of \hlight{the movie featuring Dominick Cobb as the main character} is''}$ \\
    one-hop prompt & $\mprompt = \templatey{\name{\enty}}$ & $\templatefory_\text{director}(\name{\text{Inception}}) = \text{``The name of the director of \hhlight{Inception} is''}$ \\
    \midrule
    fact composition type & \ftype{} & \catstr{stock exchange of video game's developer} \\
    first fact triplet & $(\entx, \relx, \enty)$ & (Assassin's Creed: Lost Legacy, developer, Ubisoft) \\
    second fact triplet & $(\enty, \rely, \entz)$ & (Ubisoft, stock exchange, Euronext Paris) \\\gcmidrule(lr){1-3}
    mention-constructing template & \templatex{\cdot} & $\templateforx_\text{developer}({\cdot}) = \text{``the developer of the game `$\cdots$'''}$ \\   
    prompt-constructing template & \templatey{\cdot} & $\templatefory_\text{stock~exchange}({\cdot}) = \text{``$\cdots$ is listed on a stock exchange named''}$ \\\gcmidrule(lr){1-3}
    descriptive mention of \enty{} & $\mt{} = \templatex{\name{\entx}}$ & $\templateforx_\text{developer}(\name{\text{Assassin's Creed: Lost Legacy}}) = \text{\hlight{``the developer of the game `Assassin's Creed: Lost Legacy'''}}$ \\
    two-hop prompt & $\cprompt = \templatey{\templatex{\name{\entx}}}$ & $\templatefory_\text{stock~exchange}(\templateforx_\text{developer}(\name{\text{Assassin's Creed: Lost Legacy}}))$ \\
     &  & $= \text{``\hlight{The developer of the game 'Assassin's Creed: Lost Legacy'} is listed on a stock exchange named''}$ \\
    one-hop prompt & $\mprompt = \templatey{\name{\enty}}$ & $\templatefory_\text{stock~exchange}(\name{\text{Ubisoft}}) = \text{``\hhlight{Ubisoft} is listed on a stock exchange named''}$ \\
    \bottomrule
    \end{tabular}
}
\caption{Examples from \datasetname{}. The name of the bridge entity \name{\enty{}} is shown in \hhlight{brown} font, and a descriptive mention of the bridge entity \mt{} constructed with \templatex{\name{\entx}} is shown in \hlight{purple} font.
}
\label{tab:data-examples}
\end{table*}

\section{Dataset construction}\label{sec:dataset-construction}
We construct \datasetname{} using Wikidata ~\citep{vrandevcic2014wikidata} with the following data construction pipeline.

\begin{table*}[t!]
\centering
\resizebox{0.99\textwidth}{!}{\begin{tabular}{lp{14.5cm}llrr}
\toprule
Fact Composition Type & Two-Hop Prompt \cpromptshort{} & Bridge Entity \enty{} & \entz{} & Count & Percentage \\
\midrule
actor of movie's mainchar & \hlight{The main character of the movie Dream of the Red Chamber, Part 1} was played by an actor named & Lin Daiyu & Tao Huimin & 73 & 0.16 \\
anthem of capital's cntry & The national anthem of \hlight{the country with Zagreb as its capital} is named & Croatia & Lijepa naša domovino & 204 & 0.45 \\
anthem of president's cntry & The national anthem of \hlight{the country led by president Lazarus Chakwera} is named & Malawi & Mulungu dalitsa Malaŵi & 50 & 0.11 \\
author of mainchar's novel & \hlight{The novel with 'Shere Khan' as the main character} was written by an author named & The Jungle Book & Rudyard Kipling & 308 & 0.68 \\
birthcity of cntry's president & \hlight{The president of South Korea} was born in the city of & Moon Jae-in & Geoje & 36 & 0.08 \\
birthcity of novel's author & \hlight{The author of the novel Hadrian the Seventh} was born in the city of & Frederick Rolfe & London & 3,379 & 7.41 \\
birthcity of orgz's ceo & \hlight{The CEO of Moderna} was born in the city of & Stéphane Bancel & Marseille & 189 & 0.41 \\
birthcity of person's spouse & \hlight{The spouse of Hiromi Suzuki} was born in the city of & Koji Ito & Kobe & 2,376 & 5.21 \\
birthcity of song's singer & \hlight{The singer of 'Rêver'} was born in the city of & Mylène Farmer & Pierrefonds & 1,453 & 3.19 \\
birthcntry of cntry's president & \hlight{The president of Somalia} was born in the country of & Mohamed Abdullahi Mohamed & Somalia & 36 & 0.08 \\
birthcntry of novel's author & \hlight{The author of the novel Christine} was born in the country of & Stephen King & United States of America & 3,358 & 7.36 \\
birthcntry of orgz's ceo & \hlight{The CEO of X} was born in the country of & Parag Agrawal & India & 189 & 0.41 \\
birthcntry of person's spouse & \hlight{The spouse of Vladimir Pyshnenko} was born in the country of & Natalya Meshcheryakova & Russia & 2,382 & 5.22 \\
birthcntry of song's singer & \hlight{The singer of 'Let's Get It In'} was born in the country of & Lloyd & United States of America & 1,434 & 3.15 \\
capital of anthem's cntry & The capital of \hlight{the country with the national anthem 'Fatshe leno la rona'} is & Botswana & Gaborone & 131 & 0.29 \\
capital of president's cntry & The capital of \hlight{the country led by president Ali Bongo Ondimba} is & Gabon & Libreville & 47 & 0.10 \\
cntry of person's birthcity & \hlight{The city where Aleksandăr Nikolov was born} is in the country of & Tours & France & 2,751 & 6.03 \\
cntry of univ's hqcity & \hlight{The city where the headquarters of Aichi Shukutoku University is located} is in the country of & Nagakute & Japan & 1,499 & 3.29 \\
creator of novel's mainchar & \hlight{The main character of the novel I Capture the Castle} was created by & Cassandra Mortmain & Dodie Smith & 141 & 0.31 \\
director of mainchar's movie & The name of the director of \hlight{the movie featuring Golden harp as the main character} is & Mickey and the Beanstalk & Hamilton Luske & 94 & 0.21 \\
father of novel's author & The father of \hlight{the author of the novel The Tale of Two Bad Mice} is named & Beatrix Potter & Rupert William Potter & 2,026 & 4.44 \\
father of orgz's ceo & The father of \hlight{the CEO of HarperCollins UK} is named & Charles Redmayne & Richard Charles Tunstall Redmayne & 49 & 0.11 \\
father of person's spouse & The father of \hlight{the spouse of Elsa Zylberstein} is named & Nicolas Bedos & Guy Bedos & 421 & 0.92 \\
father of song's singer & The father of \hlight{the singer of 'Étienne'} is named & Guesch Patti & Jean Porrasse & 602 & 1.32 \\
founder of ceo's orgz & \hlight{The organization led by CEO Vasily Levanov} was founded by the person named & Visual Organization & Vasily Levanov & 164 & 0.36 \\
founder of person's uguniv & \hlight{John Tien's undergrad university} was founded by the person named & United States Military Academy & Thomas Jefferson & 1,122 & 2.46 \\
founder of vdgame's dev & \hlight{The developer of the game 'Armour-Geddon'} was founded by the person named & SCE Studio Liverpool & Ian Hetherington & 3,503 & 7.68 \\
hqcity of ceo's orgz & \hlight{The organization led by CEO John Perry} has its headquarters in the city of & Bluefin Payment Systems LLC & Atlanta & 306 & 0.67 \\
hqcity of founder's dev & \hlight{The company founded by Stephen B. Streater} has its headquarters in the city of & Eidos Interactive & London & 406 & 0.89 \\
hqcity of founder's univ & \hlight{The university founded by John Wilson} has its headquarters in the city of & University of Mumbai & Mumbai & 93 & 0.20 \\
hqcity of person's uguniv & \hlight{Retta's undergrad university} has its headquarters in the city of & Duke University & Durham & 1,811 & 3.97 \\
hqcity of vdgame's dev & \hlight{The developer of the game 'The House of Da Vinci'} has its headquarters in the city of & Blue Brain Games & Bratislava & 2,310 & 5.07 \\
hqcntry of ceo's orgz & \hlight{The organization led by CEO Ties Carlier} has its headquarters in the country of & VanMoof & Netherlands & 525 & 1.15 \\
hqcntry of founder's dev & \hlight{The company founded by Anne-Laure Fanise} has its headquarters in the country of & DigixArt & France & 537 & 1.18 \\
hqcntry of founder's univ & \hlight{The university founded by Joseph Chamberlain} has its headquarters in the country of & University of Birmingham & United Kingdom & 94 & 0.21 \\
hqcntry of person's uguniv & \hlight{D. L. Waidelich's undergrad university} has its headquarters in the country of & Lehigh University & United States of America & 1,815 & 3.98 \\
hqcntry of vdgame's dev & \hlight{The developer of the game 'Terroir'} has its headquarters in the country of & General Interactive Co. & Singapore & 3,761 & 8.25 \\
mother of novel's author & The mother of \hlight{the author of the novel The Heat of the Day} is named & Elizabeth Bowen & Florence Isabella Pomeroy Colley & 1,443 & 3.16 \\
mother of person's spouse & The mother of \hlight{the spouse of Malaika Arora} is named & Arjun Kapoor & Mona Shourie Kapoor & 238 & 0.52 \\
mother of song's singer & The mother of \hlight{the singer of 'I Wanna Be Down'} is named & Brandy & Sonja Norwood & 533 & 1.17 \\
origcntry of mainchar's movie & \hlight{The movie featuring Juliane Klein as the main character} was released in the country of & Marianne and Juliane & Germany & 102 & 0.22 \\
president of anthem's cntry & \hlight{The country with the national anthem 'Azat u ankakh Artsakh'} is led by president & Republic of Artsakh & Arayik Harutyunyan & 38 & 0.08 \\
president of capital's cntry & \hlight{The country with Warsaw as its capital} is led by president & Poland & Andrzej Duda & 55 & 0.12 \\
spouse of cntry's president & The spouse of \hlight{the president of Ivory Coast} is named & Alassane Ouattara & Dominique Folloroux-Ouattara & 33 & 0.07 \\
spouse of novel's author & The spouse of \hlight{the author of the novel The Train Was on Time} is named & Heinrich Böll & Annemarie Böll & 1,597 & 3.50 \\
spouse of orgz's ceo & The spouse of \hlight{the CEO of Tethys} is named & Jean-Pierre Meyers & Françoise Bettencourt Meyers & 31 & 0.07 \\
spouse of song's singer & The spouse of \hlight{the singer of 'Last Night'} is named & Snoop Dogg & Shante & 407 & 0.89 \\
stockexch of ceo's orgz & \hlight{The organization led by CEO Luis von Ahn} is listed on a stock exchange named & Duolingo & Nasdaq & 74 & 0.16 \\
stockexch of founder's dev & \hlight{The company founded by Hae-Jin Lee} is listed on a stock exchange named & Naver Corporation & Korean Stock Exchange & 48 & 0.11 \\
stockexch of vdgame's dev & \hlight{The developer of the game 'Strider'} is listed on a stock exchange named & Capcom & Tokyo Stock Exchange & 946 & 2.07 \\
ugmajor of novel's author & In college, \hlight{the author of the novel The Masks of God} majored in & Joseph Campbell & English literature & 92 & 0.20 \\
uguniv of novel's author & As an undergrad, \hlight{the author of the novel Aiiieeeee! An Anthology of Asian-American Writers} attended the university named & Shawn Wong & University of California, Berkeley & 283 & 0.62 \\
\bottomrule
\end{tabular}}
\caption{Count of two-hop prompts for each fact composition type with examples. The text in \hlight{purple} indicates the descriptive mention \mt{} of the bridge entity. One-hop prompts \mpromptshort{} are constructed by replacing the descriptive mention with the bridge entity's name. The expanded forms of the abbreviations used for the fact composition types are listed in Table~\ref{tab:abbr}.}
\label{tab:dataset}
\end{table*}

\subsection{Data Selection}\label{sec:data-selection}
We select relations and entities that are well-known and result in sufficient numbers of samples per relation. Relations are selected manually. At the time of querying Wikidata, we constrain entities to singular entities with natural language Wikipedia titles and select entities with a maximal number of reference links. We also exclude the cases of $\entx = \enty$ that might allow trivial recall of \enty{} by directly copying from the input. In addition, we make sure that bridge entities \enty{} are unique among the facts of the same fact composition type to mitigate the imbalance in the bridge entity. Finally, we apply down-sampling to mitigate the imbalance in the fact composition type.

\paragraph{Relation Selection}
First, we determine the type of the bridge entity's descriptive mention by selecting the type of entities \entx{} and relation \relx{} to collect $\outx{} = \enty{}$. The bridge entities we select have types like \catstr{song's singer} (the singer of a specific song), \catstr{country's anthem} (the country with a specific national anthem), \catstr{founder's organization} (the organization founded by a specific person), and \catstr{organization's ceo} (the CEO of a specific organization). For example, while there can be many authors for some novels, \catstr{author's novel} is selected as a type of descriptive mention of the bridge entity because we can use only the novels with a single author. We determine 19 types of bridge entity's descriptive mention with this process.

Now that we have \btype{} determined, we determine the type of relations \rely{} to determine the type of the fact composition, \ftype{}. Note that \btype{} determined in the previous step falls into the category of country, organization (organization, undergraduate university, game developer), real person (author, president, CEO, spouse, singer), fictional character (main character), movie, novel, or city (headquarters city). Note that \btype{} is also the bridge entity itself that the descriptive mention refers to. Therefore, we select \rely{} that are likely to give us a sufficient number of $(\enty, \rely, \entz)$ where \entz{} is the only object entity satisfying the relation \rely{} for these categories of \enty{}. As in the previous step, we select common relations as \rely{}. Using the selected types of \rely{}, we create 52 fact composition types including \catstr{mother of song's singer} (the city where the novel of a specific novel was born), \catstr{headquarterscity of video game's developer} (the city where the headquarters of the developer of a specific video game is located), and \catstr{director of main character's movie} (the director of the movie which has a specific character as the main character).

\paragraph{Querying Wikidata}
We collect the fact triplets of the selected fact composition types through Wikidata Query Service\footnote{\href{https://query.wikidata.org}{https://query.wikidata.org}} with one handcrafted query for each of the 52 fact composition types. When there are too many results for the API call to bring before a timeout occurs, we reduce the number of the results by filtering the results with the number of reference links and/or adding other conditions to the query.

For the relations that are subject to change by nature, e.g., CEO of a company, we retrieve the information at the time of January 1, 2022. We choose this timestamp considering the training time of LLaMA-2~\citep{Touvron2023-jd} models that we use for our study. To analyze LLMs trained at different times, filtering the dataset with model accuracy before the analysis or using the prompts of the fact composition types that are less likely to change over time (e.g., ``founder of videogame's developer'') would resolve potential issues from the temporality of the dataset.

\subsection{Natural Language Templates}\label{sec:natural-language-templates}
We manually create natural language templates. To this end, we first create descriptive mentions of the bridge entity. To create the descriptive mentions, we manually write \relx{}-specific \textit{mention-constructing templates} \templatex{\cdot}. For example, $\templateforx_\text{singer}(\cdot) = \text{``the singer of `$\cdots$'''}$ creates $\mt = \text{``the singer of `Superstition'''}$.

Next, we create one/two-hop prompt templates. We manually write \rely{}-specific \textit{prompt-constructing templates} \templatey{\cdot} that take a mention of the bridge entity \enty{} and form a prompt querying about \enty{}'s relational attribute \rely{} in a way that the prompt can be correctly answered with a mention of \entz{}. For example, $\templatefory_\text{mother}(\cdot) =\ $``The mother of $\cdots$ is'' is used to create the one-hop prompt ``The mother of Stevie Wonder is'' and also the two-hop prompt ``The mother of the singer of `Superstition' is''.

We write one representative template for each \templatefx{} and \templatefy{} in a way that two-hop prompts are natural. Some examples of how the templates are used to construct the prompts are shown in Table~\ref{tab:data-examples}. Afterward, we translate the collected fact triplets to pairs of two-hop prompts and one-hop prompts using the manually written templates. To represent entities in a string, we use the title of the entity's Wikidata page. We ensure that the generated prompts are grammatically correct. Table~\ref{tab:dataset} shows the actual examples of the two-hop prompts and the bridge entity for each fact composition type.

\begin{table}[t!]
\centering
\resizebox{0.25\textwidth}{!}{\begin{tabular}{ll}
\toprule
Abbreviation & Full Term \\
\midrule
hq & headquarters \\
ug & undergrad \\
orig & origin \\
univ & university \\
stockexch & stock exchange \\
orgz & organization \\
mainchar & main character \\
vdgame & videogame \\
cntry & country \\
dev & developer \\
\bottomrule
\end{tabular}}
\caption{Abbreviations used for the fact composition types.}
\label{tab:abbr}
\end{table}

\begin{figure}[t!]
\centering
\includegraphics[width=.48\textwidth]{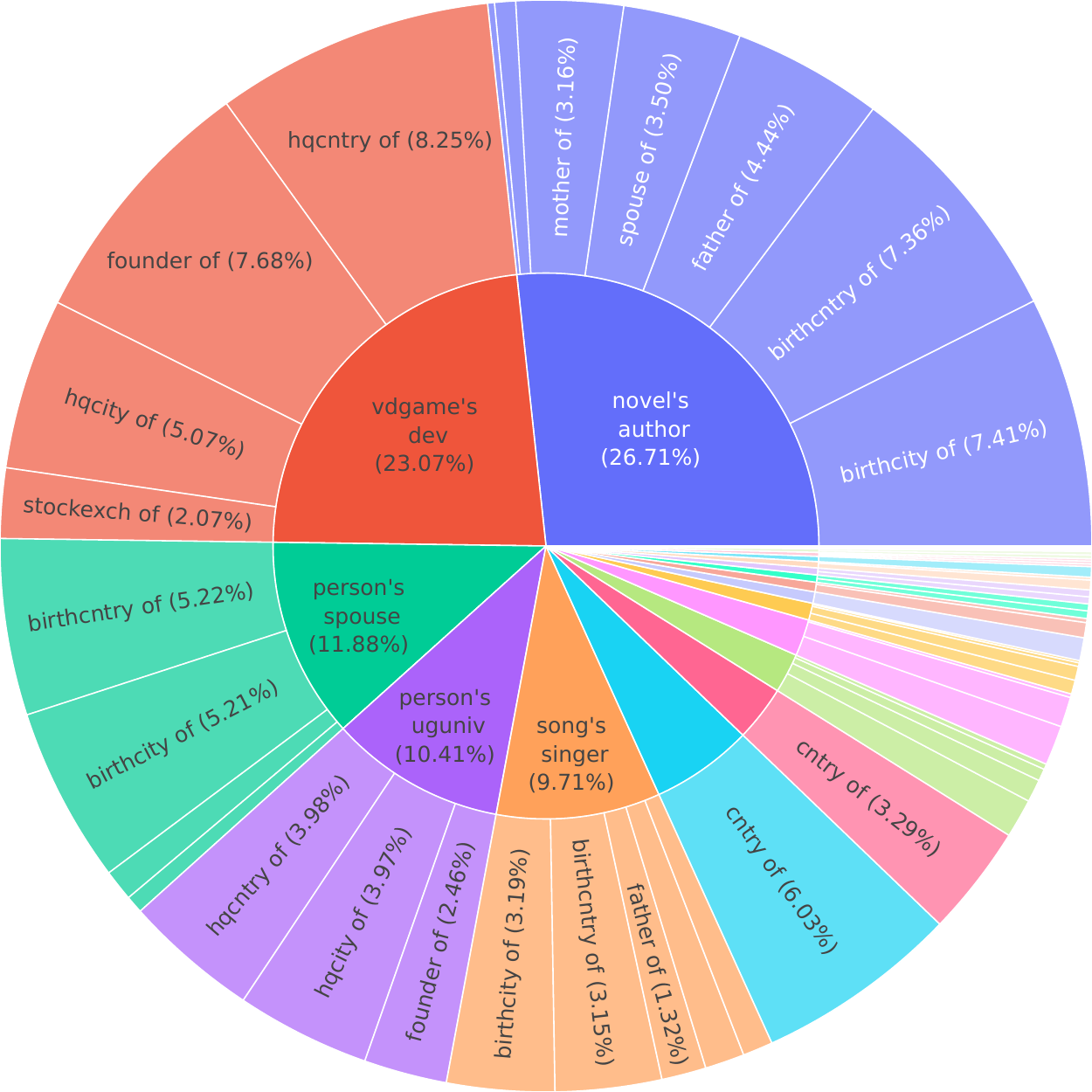}
\caption{Statistics of the dataset of \datasetname{}. The inner part shows the percentage of two-hop prompts with the type of descriptive mention of the bridge entity: \btype{}. The outer part shows the percentage of the two-hop prompts with the fact composition type: \ftype{} (only \text{\cat{\rely{}} of} is shown as the annotation) in \datasetname{}. The expanded forms of the abbreviations used for the fact composition types are listed in Table~\ref{tab:abbr}.}\label{fig:sunburst}
\end{figure}

\section{Dataset Statistics}\label{sec:dataset-stats}

\datasetname{} consists of 45,595 unique pairs of fact triplets $((\entx, \relx, \enty), (\enty, \rely, \entz))$ of 52 fact composition types, translated into 45,595 one/two-hop prompts.
Figure~\ref{fig:sunburst} shows the distribution of the fact composition types. The distribution of the fact composition type is relatively balanced, with the type that has the largest portion covering only 7.41\% of the dataset (\catstr{birth city of novel's author}).

\begin{figure*}[t!]
  \centering
  \begin{subfigure}[b]{.50\textwidth}
  \includegraphics[width=\textwidth]{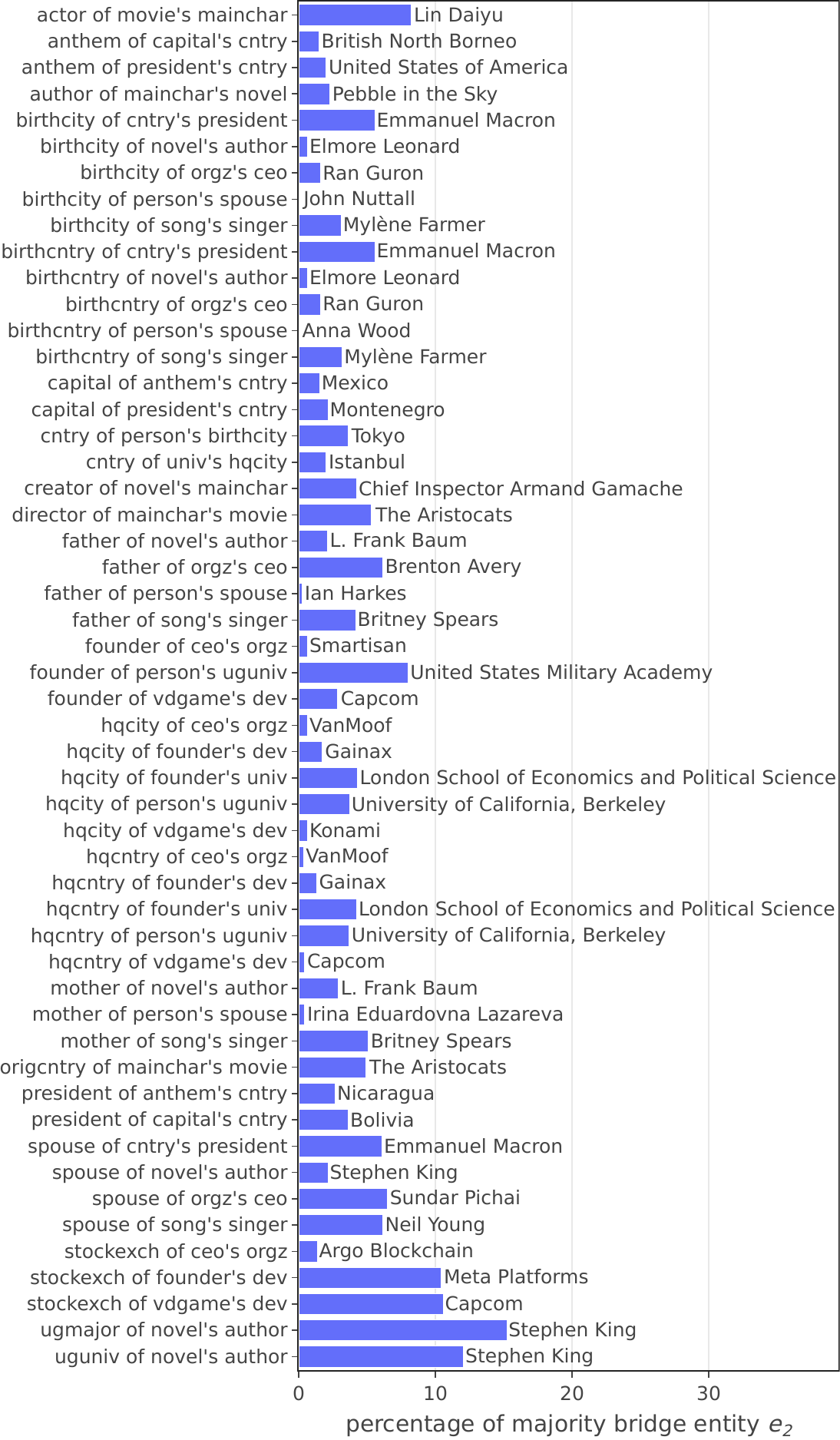}
  \caption{Percentage of majority bridge entity \enty{}}
  \label{fig:majority-entity}
  \end{subfigure}\hfill
  \begin{subfigure}[b]{.43\textwidth}
  \includegraphics[width=\textwidth]{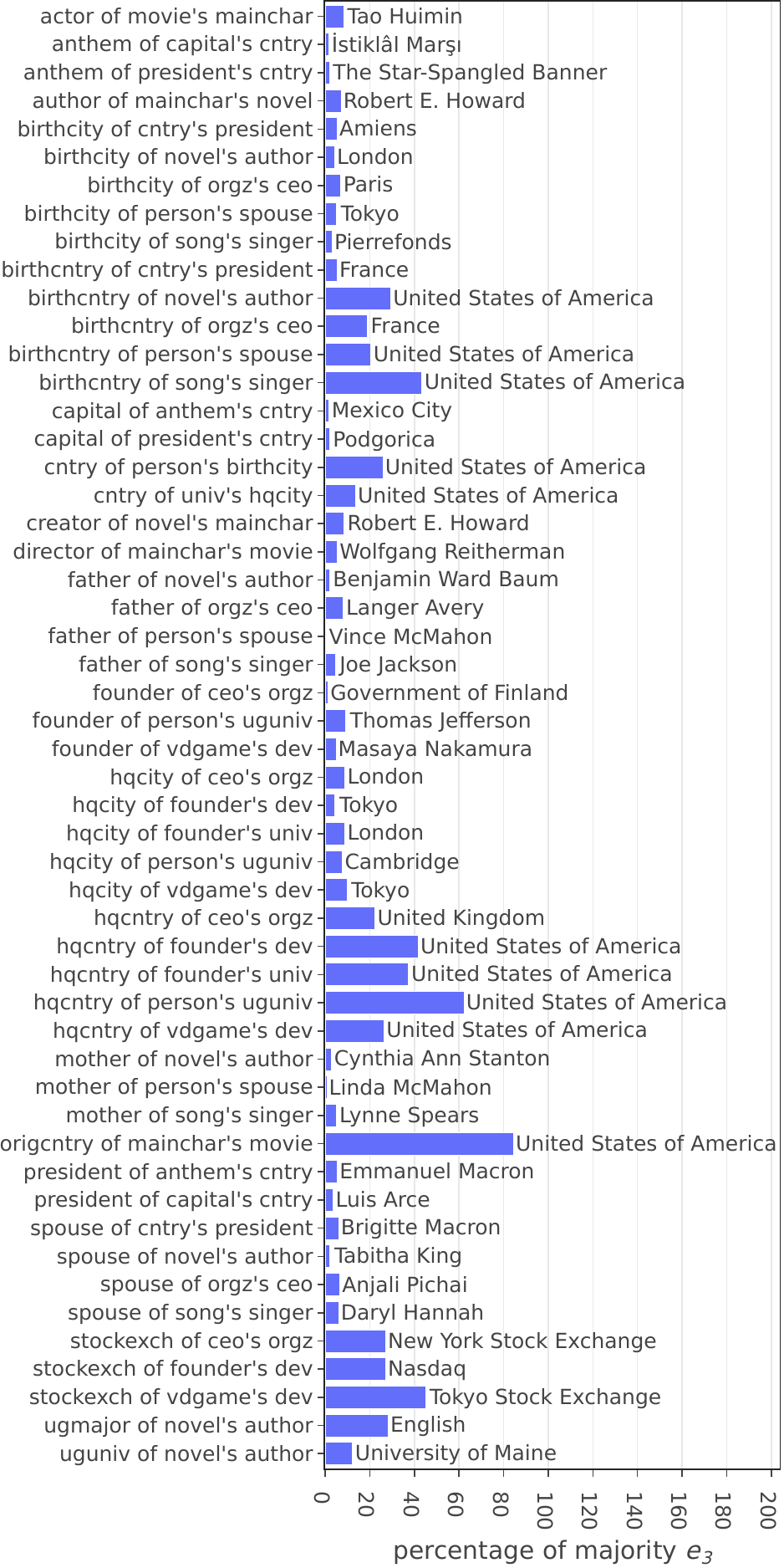}
  \caption{Percentage of majority \entz{}}
  \label{fig:majority-answer}
  \end{subfigure}
\caption{Percentage of the most frequent entities for each fact composition type of \datasetname{}. The expanded forms of the abbreviations used for the fact composition types are listed in Table~\ref{tab:abbr}.}
\end{figure*}

Figure~\ref{fig:majority-entity} shows the percentage of the majority bridge entity \enty{}, i.e., \enty{} that is utilized the most to construct the one-hop prompt that corresponds to each two-hop prompt. The highest percentage of majority bridge entity among all fact composition types is only 15\%, showing that the dataset is not biased as favorable towards certain \enty{}. Figure~\ref{fig:majority-answer} shows the percentage of majority \entz{} that serve as the ground truth answer for the two-hop prompts. Table~\ref{tab:dataset} shows the number of two-hop prompts for each fact composition type with examples. We ensure that the number of prompts for a fact composition type exceeds at least 30 for statistically significant results.

\section{Justification of Internal Entity Recall Score: Appositive Generation Experiment}\label{sec:apposition}
\begin{figure}[ht!]
  \centering
  \includegraphics[width=0.4\textwidth]{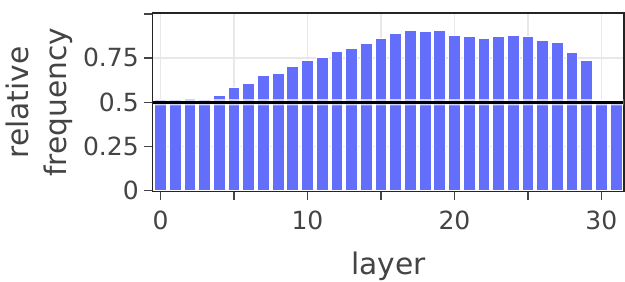}
  \caption{The relative frequency of the cases where increasing the entity recall score at a layer increases the probability of the model to output \mentyfirst{} as the next token of a comma following the prefix of \cpromptshort{} ending at the descriptive mention (\textit{``The mother of the singer of `Superstition',''}), for LLaMA-2 7B.}
  \label{fig:apposition}
\end{figure}

\paragraph{Experiment}
We demonstrate that \xmetricop{} is a reasonable approximation of the internal recall of the bridge entity with indirect evidence.
Note that \xmetricreal{l} is calculated not at the last token of \cpromptshort{} but at the last token of the bridge entity's descriptive mention, where it is grammatically natural to prepend a comma followed by the name of \menty{} (e.g., ``The mother of the singer of `Superstition'\textit{, Stevie Wonder}''). In the resulting string, grammatically \mt{} becomes the \textit{antecedent} and \menty{} becomes the \textit{appositive}; an appositive is a noun phrase that follows another noun phrase in opposition to it and provides information that further identifies or defines it, and the antecedent is the noun phrase that the appositive describes. Then, if \xmetricreal{l} reasonably approximates the internal recall of the bridge entity \enty{}, it is expected that \textit{there will be at least some layers $l$ where increasing \xmetricreal{l} increases the relative frequency of the LLM to generate \appositiontarget{} with a relative frequency higher than random chance}. In other words, we check the relative frequency of the cases where increasing the entity recall score at a layer increases the probability of the model to output \mentyfirst{} as the next token of a comma following the prefix of \cpromptshort{} ending at the descriptive mention (\textit{``The mother of the singer of `Superstition',''}). We calculate this relative frequency as described in Section~\ref{sec:rq2-setup} but using the probability instead of \ymetricop{}.

\begin{figure*}[t!]
  \centering
  \includegraphics[width=1.0\textwidth]{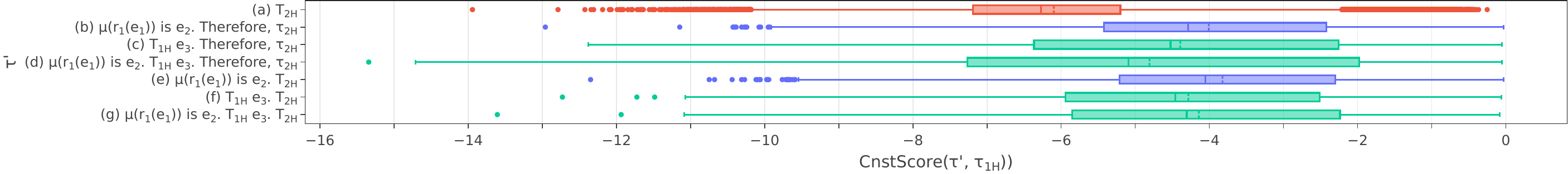}
  \caption{Distribution of \ymetricop{} calculated for different styles of prompts $\tau'$ for LLaMA-2 7B.}
  \label{fig:main-ymetric}
\end{figure*}

\paragraph{Result}
Figure~\ref{fig:apposition} demonstrates that, in most of the mid-late layers, increasing the latent recall of the bridge entity when the LLM processes \mt{} also increases the relative frequency of the LLM to output \mentyfirst{} to generate the appositive of \mt{} followed by a comma.\footnote{For this analysis, we exclude the cases where the descriptive mention ends with one of the following: ?', .', !', ,', '', )', ''', where appending a comma introduces changes in the tokenization results for LLaMA-2.} The result indicates that \xmetricop{} at the $n$-th token has controllability of the token to be generated as the $n+2$-th token to make it more likely to be the first token of the appositive, serving as an indirect evidence that \xmetricreal{l} is a reasonable proxy of the internal recall of the bridge entity.

\section{Justification of Consistency Score: Comparative Experiment with Chain-of-Thought Cases}\label{sec:cot}
\paragraph{Experiment}
We demonstrate that the proposed definition of \ymetricreal{} is a reasonable proxy of the utilization of what the LLM knows about the bridge entity's attribute -- the latent recall of its answer to \mpromptshort{} -- with indirect evidence. If the information to reason with is given as part of the input, e.g., if the given prompt is \textit{``The singer of `Superstition' is Stevie Wonder. The mother of Stevie Wonder is named Lula. The mother of the singer of `Superstition' is''}, the LLM would not need to internally perform the multi-hop reasoning to refer to what its output to the one-hop prompt \textit{``The mother of Stevie Wonder is''} is, but just copy the answer from the input. Therefore, \ymetricop{} of such a case will be lower than the case where the LLM needs to internally figure out what its answer to the one-hop prompt given the hint of who the descriptive mention refers to, e.g., \textit{``The singer of `Superstition' is Stevie Wonder. The mother of the singer of `Superstition' is''}. Therefore, to check whether this is the case, we compare \ymetricop{} computed with the several Chain-of-Thought (CoT) style prompts $\tau'$, i.e., \ $\ymetricop(\tau', \mpromptshort)$.

\paragraph{Result}
Figure~\ref{fig:main-ymetric} shows the distribution of \ymetricop{} computed with different styles of prompts $\tau'$ as written in the y-axis. The red case is the consistency score of the two-hop prompt that we mainly study in our work, which requires full multi-hop reasoning. Because no information to reason from is given in the input, \ymetricop{} is significantly lower than the cases of other CoT-style prompts. The blue case is where what the descriptive mention refers to is given as the input, but what the LLM knows about the bridge entity's attribute needs to be internally recalled and referred to. The green cases are where the bridge entity's attribute, i.e., the answer to the prompt, is explicitly given in the input, and thus, the LLM does not need to refer to its answer to the one-hop prompt. The result demonstrates that the mean of \ymetricop{} is higher for the blue cases where the model is forced to refer to its answer to the one-hop prompt than in the green cases where the model does not need to refer to the answer. The difference between the red and the blue cases would have come from the existence of the information of the descriptive mention's identity in the input prompt, which would have helped the LLM to use the connection to refer to what it knows about the bridge entity.

\begin{table*}[t!]
\centering
\resizebox{0.99\textwidth}{!}{\begin{tabular}{lllll}
\toprule
Descriptive Mention Type & 0 & 1 & 2 & 3 \\
\midrule
novel's author & a critic of \hlight{\name{\entx}} & the filmmaker of \hlight{\name{\entx}} & the main character of \hlight{\name{\entx}} & a fan of \hlight{\name{\entx}} \\
person's birth city & the city where \hlight{\name{\entx}} never visited & the city where \hlight{\name{\entx}} is abandoned & the city where \hlight{\name{\entx}} is banned & the city where \hlight{\name{\entx}} never lived in \\
orgz's ceo & the COO of \hlight{\name{\entx}} & the rival of \hlight{\name{\entx}} & the CTO of \hlight{\name{\entx}} & the CFO of \hlight{\name{\entx}} \\
capital's cntry & the country which does not have \hlight{\name{\entx}} as its city & the country which does not have \hlight{\name{\entx}} as its capital & the country which does not have \hlight{\name{\entx}} as its largest city &  \\
president's cntry & the country where \hlight{\name{\entx}} is not the president & the country where \hlight{\name{\entx}} is not the head of state & the country where \hlight{\name{\entx}} is a rival of the president & the country where \hlight{\name{\entx}} is a critic of the president \\
anthem's cntry & the country which does not have \hlight{\name{\entx}} as its anthem & the country which banned singing \hlight{\name{\entx}} & the country where \hlight{\name{\entx}} is blacklisted & the country where \hlight{\name{\entx}} is banned \\
vdgame's dev & a competitor of \hlight{\name{\entx}} & a plagiarist of \hlight{\name{\entx}} & a critic of \hlight{\name{\entx}} & a rival of \hlight{\name{\entx}} \\
founder's dev & the company \hlight{\name{\entx}} criticizes & a critic of \hlight{\name{\entx}} & a competitor to \hlight{\name{\entx}} & the company \hlight{\name{\entx}} is a rival of \\
univ's hqcity & the city where \hlight{\name{\entx}} is not located & the city where \hlight{\name{\entx}} is not headquartered & the city where \hlight{\name{\entx}} is not founded & the city where \hlight{\name{\entx}} is not established \\
movie's mainchar & the antagonist in \hlight{\name{\entx}} & a sidekick in \hlight{\name{\entx}} & an extra in \hlight{\name{\entx}} & a critic of \hlight{\name{\entx}} \\
novel's mainchar & the antagonist in \hlight{\name{\entx}} & a sidekick in \hlight{\name{\entx}} & an extra in \hlight{\name{\entx}} & a critic of \hlight{\name{\entx}} \\
mainchar's novel & the novel where \hlight{\name{\entx}} is not the main character & the novel where \hlight{\name{\entx}} does not appear & the novel where \hlight{\name{\entx}} is not the protagonist & the novel where \hlight{\name{\entx}} is not the antagonist \\
mainchar's movie & the movie where \hlight{\name{\entx}} is not the main character & the movie where \hlight{\name{\entx}} does not appear & the movie where \hlight{\name{\entx}} is not the protagonist & the movie where \hlight{\name{\entx}} is not the antagonist \\
ceo's orgz & the company \hlight{\name{\entx}} criticizes & a critic of \hlight{\name{\entx}} & a competitor to \hlight{\name{\entx}} & the company \hlight{\name{\entx}} is a rival of \\
cntry's president & a critic of \hlight{\name{\entx}} & a protester against \hlight{\name{\entx}} & a rival of \hlight{\name{\entx}} & a competitor to \hlight{\name{\entx}} \\
song's singer & a critic of \hlight{\name{\entx}} & a singer covering \hlight{\name{\entx}} without permission & a plagiarist of \hlight{\name{\entx}} & a rival of \hlight{\name{\entx}} \\
person's spouse & the father of \hlight{\name{\entx}} & the mother of \hlight{\name{\entx}} & a child of \hlight{\name{\entx}} & a sibling of \hlight{\name{\entx}} \\
person's uguniv & the university where the application of \hlight{\name{\entx}} was rejected & the university where \hlight{\name{\entx}} never went to & the university where \hlight{\name{\entx}} was not accepted & the university where \hlight{\name{\entx}} was not admitted \\
founder's univ & the university where \hlight{\name{\entx}} graduated from & the alma mater of \hlight{\name{\entx}} & the university where \hlight{\name{\entx}} was admitted to & the university where \hlight{\name{\entx}} was accepted to \\
\bottomrule
\end{tabular}}
\caption{Candidate templates of \negrelx{} for each type of descriptive mention of the bridge entity. The expanded forms of the abbreviations used for the fact composition types are listed in Table~\ref{tab:abbr}.}
\label{tab:negrelx}
\end{table*}

\section{Technical Details}\label{sec:technical-details}
We modify the codebase of \citet{nanda2022transformerlens} to run the experiments before refactoring. We use 1-8 40GB A100 GPUs for the experiments. All experiments run in less than 24 hours. We use the model weights from HuggingFace Transformers~\citep{wolf2020huggingfaces} and use full precision for LLaMA-2 7B and 13B and half-precision for 70B. The SPARQL queries for querying Wikidata are written with the help of GPT-4~\citep{openai2023gpt4}.

\section{Accuracy-based Analysis for the Second Hop of Multi-Hop Reasoning}\label{sec:accuracy}

\begin{figure*}[t!]
\centering
    \begin{subfigure}[t]{0.32\textwidth}
    \includegraphics[width=\textwidth]{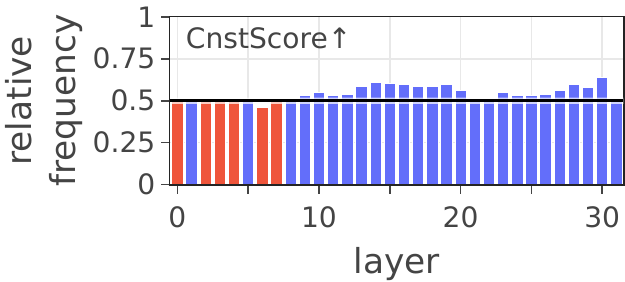}
    \caption{LLaMA-2 7B with \ymetricop{}}\label{fig:correct_const_7b}
    \end{subfigure}
    \hfill
    \begin{subfigure}[t]{0.32\textwidth}
    \includegraphics[width=\textwidth]{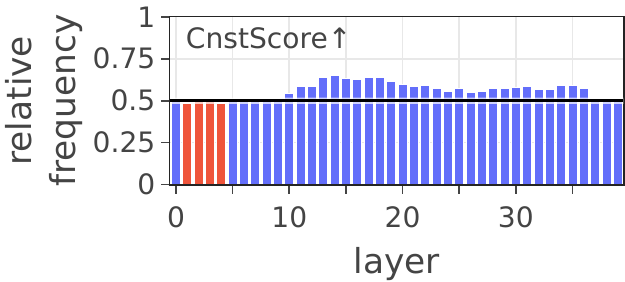}
    \caption{LLaMA-2 13B with \ymetricop{}}\label{fig:correct_const_13b}
    \end{subfigure}
    \hfill
    \begin{subfigure}[t]{0.32\textwidth}
    \includegraphics[width=\textwidth]{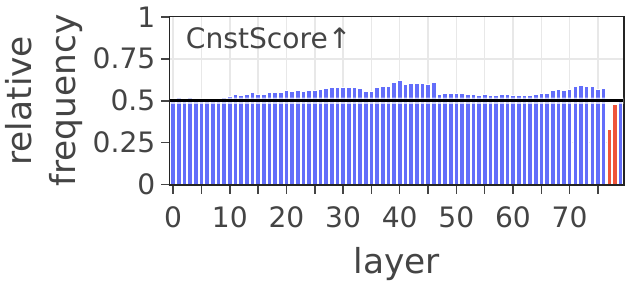}
    \caption{LLaMA-2 70B with \ymetricop{}}\label{fig:correct_const_70b}
    \end{subfigure}
    \begin{subfigure}[t]{0.32\textwidth}
    \includegraphics[width=\textwidth]{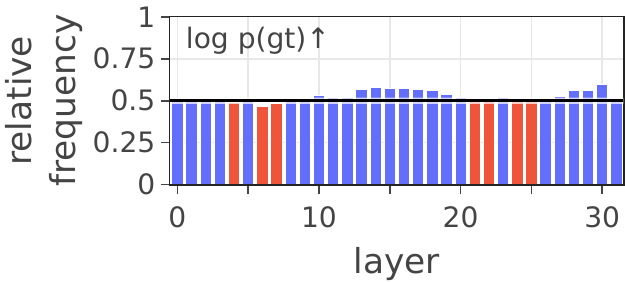}
    \caption{LLaMA-2 7B with log probability}\label{fig:correct_logp_7b}
    \end{subfigure}
    \hfill
    \begin{subfigure}[t]{0.32\textwidth}
    \includegraphics[width=\textwidth]{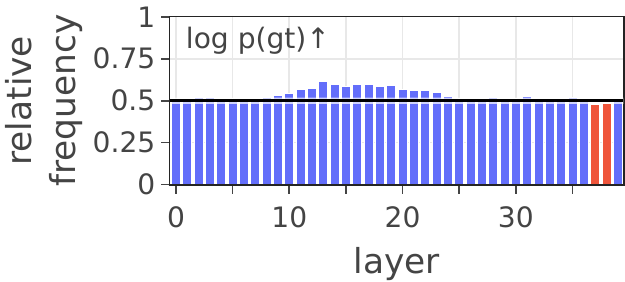}
    \caption{LLaMA-2 13B with log probability}\label{fig:correct_logp_13b}
    \end{subfigure}
    \hfill
    \begin{subfigure}[t]{0.32\textwidth}
    \includegraphics[width=\textwidth]{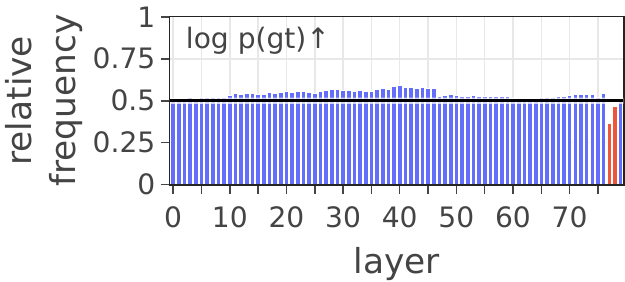}
    \caption{LLaMA-2 70B with log probability}\label{fig:correct_logp_70b}
    \end{subfigure}
\caption{Relative frequency that stronger recall of the bridge entity at the $l$-th layer increases the consistency score (top row) or the log probability of the first token of the ground truth answer (bottom row) of the LLM. The relative frequency is calculated only for the cases where the one-hop prompt is completed with the ground truth answer. Bars are colored blue if the relative frequency is greater than or equal to 0.5 and red otherwise. We manually set the value of 0.5 at the last layer because the intervention does not affect the consistency at that layer.}
\label{fig:correct}
\end{figure*}

We perform consistency-based analysis instead of an accuracy-based analysis because solely relying on the answer correctness has limitations in answering our research questions, as explained in Section~\ref{sec:rq2-metric}. For further analysis, we present accuracy-based results in this section.

\paragraph{Using log probability of the ground truth answer instead of the consistency score does not affect our findings.} We perform the RQ2 experiment described in Section~\ref{sec:rq2-setup} not with \ymetricop{}, but with the output log probability of the first token of the ground truth answer, e.g., \textit{``Lula''} for the two-hop prompt \textit{``The mother of the singer of `Superstition' is''}. We measure the relative frequency using the 17,231/45,595 two-hop prompts of which the corresponding one-hop prompts, e.g., \textit{``The mother of Stevie Wonder is''}, are completed with the ground truth answer by all of the 7B, 13B, and 70B models.

Figure~\ref{fig:correct} shows the results, where the top row contains the relative frequency with \ymetricop{} and the bottom row contains the relative frequency with the log probability of the ground truth answer. When scaling from 7B to 13B and 70B, the maximum relative frequency with \ymetricop{} is 0.64 in layer 30 (7B), 0.65 in layer 14 (13B), and 0.62 in layer 40 (70B). The maximum relative frequency with the log probability of the ground truth answer is 0.60 in layer 30 (7B), 0.62 in layer 13 (13B), and 0.59 in layer 40 (70B). While the values are slightly lower for the log probability than those for the consistency score, the overall trends are alike. Also, as also observed in Section~\ref{sec:rq2-results} with consistency, the second-hop reasoning with the log probability does not strengthen with increasing model size.

\paragraph{Filtering based on the accuracy of the one-hop prompt does not affect our findings.}

\begin{figure*}[t!]
\centering
    \begin{subfigure}[t]{0.32\textwidth}
    \includegraphics[width=\textwidth]{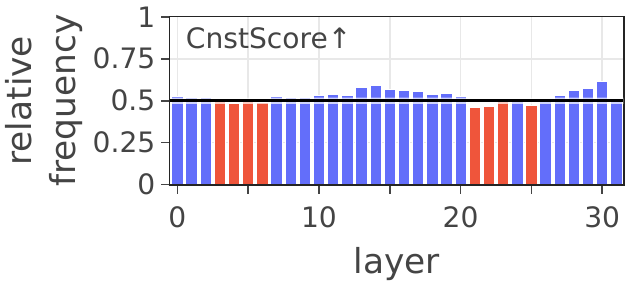}
    \caption{LLaMA-2 7B \text{\mpromptshort{} correct} cases}\label{fig:compare_const_7b_correct}
    \end{subfigure}
    \hfill
    \begin{subfigure}[t]{0.32\textwidth}
    \includegraphics[width=\textwidth]{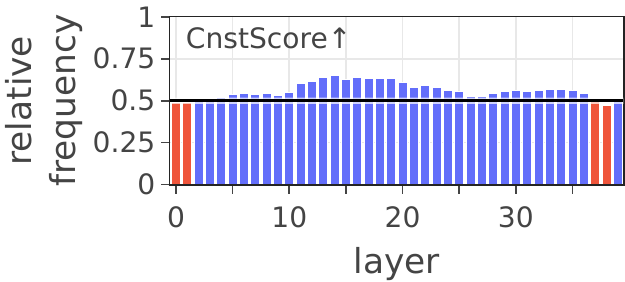}
    \caption{LLaMA-2 13B \text{\mpromptshort{} correct} cases}\label{fig:compare_const_13b_correct}
    \end{subfigure}
    \hfill
    \begin{subfigure}[t]{0.32\textwidth}
    \includegraphics[width=\textwidth]{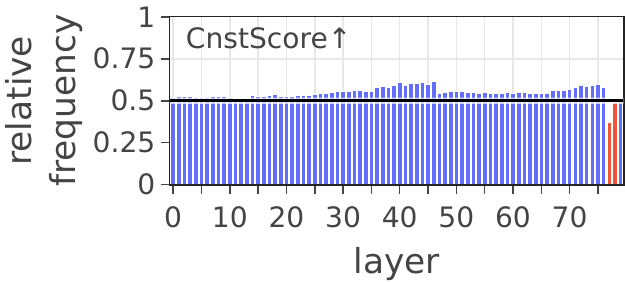}
    \caption{LLaMA-2 70B \text{\mpromptshort{} correct} cases}\label{fig:compare_const_70b_correct}
    \end{subfigure}
    \begin{subfigure}[t]{0.32\textwidth}
    \includegraphics[width=\textwidth]{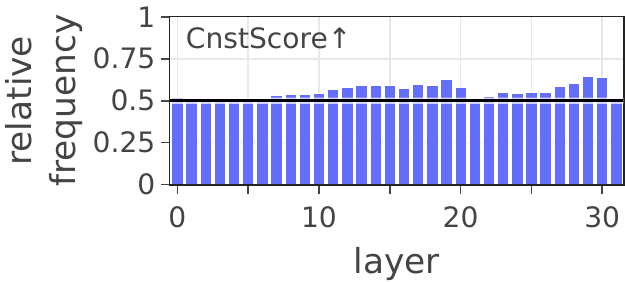}
    \caption{LLaMA-2 7B \text{\mpromptshort{} incorrect} cases}\label{fig:compare_const_7b_incorrect}
    \end{subfigure}
    \hfill
    \begin{subfigure}[t]{0.32\textwidth}
    \includegraphics[width=\textwidth]{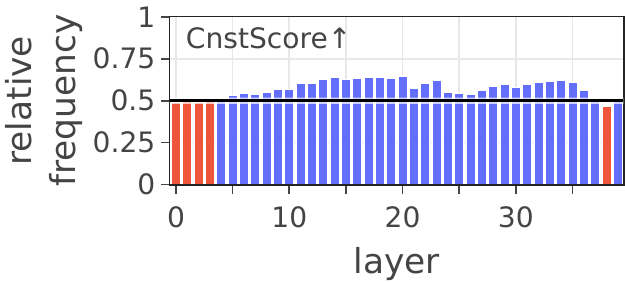}
    \caption{LLaMA-2 13B \text{\mpromptshort{} incorrect} cases}\label{fig:compare_const_13b_incorrect}
    \end{subfigure}
    \hfill
    \begin{subfigure}[t]{0.32\textwidth}
    \includegraphics[width=\textwidth]{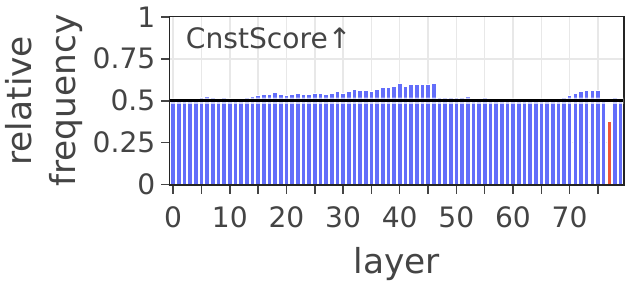}
    \caption{LLaMA-2 70B \text{\mpromptshort{} correct} cases}\label{fig:compare_const_70b_incorrect}
    \end{subfigure}
\caption{Relative frequency that stronger recall of the bridge entity at the $l$-th layer increases the consistency of the LLM, compared between the cases where the one-hop prompt is correctly completed with one of the ground truth answer candidates (\mpromptshort{} correct; top row) and not (\mpromptshort{} incorrect; bottom row). Bars are colored blue if the relative frequency is greater than or equal to 0.5 and red otherwise. We manually set the value of 0.5 at the last layer because the intervention does not affect the consistency at that layer.}
\label{fig:compare_correct_incorrect}
\end{figure*}

We test whether the second hop of the latent multi-hop reasoning is stronger for the two-hop prompts of which the corresponding one-hop prompts are completed correctly with one of the ground truth answer candidates. For this analysis, we filter the two-hop prompts into two sets: those where the corresponding one-hop prompts are correctly completed for all model scales (\text{\mpromptshort{} correct}) and those where the corresponding one-hop prompt is not completed with any of the ground truth answer candidates for all model scales (\text{\mpromptshort{} incorrect}). Since the trend of the relative frequency significantly varies for different fact composition types (Figure~\ref{fig:rq2_main}), for fair comparison, we make the distribution of the fact composition types of the two sets become the same by sampling. This results in two sets of 9,734 two-hop prompts with the same distribution of fact composition types.

Figure~\ref{fig:compare_correct_incorrect} shows that the overall trends are similar for the two sets, \text{\mpromptshort{} correct} (top row) and \text{\mpromptshort{} incorrect} (bottom row), and that the maximum relative frequency of the two sets do not differ much across different model sizes. When scaling from 7B to 13B and 70B, the maximum relative frequency for the \text{\mpromptshort{} correct} set is 0.62 in layer 30 (7B), 0.65 in layer 14 (13B), and 0.62 in layer 46 (70B). The maximum relative frequency for the \text{\mpromptshort{} incorrect} set is 0.65 in layer 29 (7B), 0.65 in layer 20 (13B), and 0.61 in layer 46 (70B).

\end{document}